\documentclass[a4paper,fleqn]{cas-dc}

\usepackage[numbers]{natbib}

\graphicspath{{images/}} %
\newcommand*\rfrac[2]{{}^{#1}\!/_{#2}}
\usepackage{subcaption}
\usepackage{listings}
\usepackage{array}

\lstdefinestyle{Python}{
    language        = Python,
    deletekeywords={in},
    basicstyle      = \ttfamily,
    keywordstyle    = \color{blue},
    stringstyle     = \color{green},
    commentstyle    = \color{red}\ttfamily,
    basicstyle      = \small,
    showstringspaces=false
}

\def\tsc#1{\csdef{#1}{\textsc{\lowercase{#1}}\xspace}}
\tsc{WGM}
\tsc{QE}
\tsc{EP}
\tsc{PMS}
\tsc{BEC}
\tsc{DE}

\begin{document}
\let\WriteBookmarks\relax
\def\floatpagepagefraction{1}
\def\textpagefraction{.001}
\shorttitle{Deep learning of pore defects in real and synthetic additive manufacturing powders}
\shortauthors{A.\ Bjerregaard et~al.}

\title [mode = title]{%
Attenuation-adjusted deep learning of pore defects in 2D radiographs of additive manufacturing powders
}

\author[1]{Andreas Bjerregaard}[orcid=0000-0003-2742-6178]
\cormark[1]

\ead{anje@di.ku.dk}

\author[2]{David Schumacher}[orcid=0000-0002-8435-4978]
\ead{david.schumacher@bam.de}

\author[1]{Jon Sporring}[orcid=0000-0003-1261-6702]
\ead{sporring@di.ku.dk}

\address[1]{Department of Computer Science, University of Copenhagen, Universitetsparken 1, DK-2100 Copenhagen, Denmark}

\address[2]{Bundesanstalt für Materialforschung und –prüfung (BAM), Unter den Eichen 87, 12205 Berlin, Germany}

\cortext[cor1]{Corresponding author}
\tnotetext[2]{Implementation found on
\url{https://github.com/yhsure/porosity}}
\begin{abstract}
The presence of gas pores in metal feedstock powder for additive manufacturing greatly affects the final AM product. Since current porosity analysis often involves lengthy X-ray computed tomography (XCT) scans with a full rotation around the sample, motivation exists to explore methods that allow for high throughput — possibly enabling in-line porosity analysis during manufacturing. Through labelling pore pixels on single 2D radiographs of powders, this work seeks to simulate such future efficient setups. High segmentation accuracy is achieved by combining a model of X-ray attenuation through particles with a variant of the widely applied UNet architecture; notably, F1-score increases by $11.4\%$ compared to the baseline UNet. The proposed pore segmentation is enabled by: 1) pretraining on synthetic data, 2) making tight particle cutouts, and 3) subtracting an ideal particle without pores generated from a distance map inspired by Lambert-Beers law. This paper explores four image processing methods, where the fastest (yet still unoptimized) segments a particle in mean $0.014\,s$ time with F1-score $0.78$, and the most accurate in $0.291\,s$ with F1-score $0.87$. Due to their scalable nature, these strategies can be involved in making high throughput porosity analysis of metal feedstock powder for additive manufacturing.
\end{abstract}

\begin{keywords}
additive manufacturing \sep non-destructive testing \sep porosity estimation \sep radiography \sep image analysis \sep image segmentation \sep computer vision 
\end{keywords}

\maketitle

\section{Introduction}
Some of the most commonly used additive manufacturing (AM) techniques to produce complex-shaped metal parts are laser powder bed fusion (LPBF) and electron beam melting (EBM). However, the mechanical integrity of the final product heavily depends on internal defects, such as pores. These local variations in density, i.e. distribution of pores, are influenced by numerous parameters \cite{Vock2019}. On the one hand, the technical parameters like hatching strategy or energy input per volume depend on the process itself and can be optimised. On the other hand, the inherent quality of the feedstock powder can severely affect the overall quality of the final product~\cite{Sola2019}. Especially the pores, which are already present in the feedstock powder, can remain in the final product and even regrow when heat treated~\cite{Cunningham2017, Tammas-Williams2016}.

When looking at final AM parts, the types of pores can be divided into different classes based on their origin. When the energy input per volume during the process is too low, the bulk material is not entirely molten which causes present cavities to remain within the product instead of diffusing out of the melt pool. In case of a too high energy input, certain components of the material start to evaporate, leaving elongated keyhole-shaped pores. Furthermore, a very high cooling rate and hence a fast solidification of the molten material can trap gas bubbles which results in mainly circular-shaped pores within the final AM part. And this is where quality control of feedstock powders steps in, since the trapped gas bubbles can originate from either moisture in the powder bed, atmospheric environment, or from inherent pores in the feedstock powder it self. Hence, it is crucial for the integrity of the final AM part to assess the quality of the feedstock powder which this study aims to contribute to.

The most relevant quality features of feedstock powders, which are controlled before and during the process, are: level of moisture, flowability, which is connected to the particle size distribution, purity of the used metal alloys and more often the relative density of the powder. The latter is very difficult to measure, especially when dealing with powder particle sizes below 100~µm. State-of-the-art methods to determine the density of feedstock powders are pycnometry or using Archimedes' principle. Those methods are comparably fast, but they do not yield distinct pore size distributions or even pore locations. 

These limitations can be overcome when applying X-ray computed tomography (XCT) which is capable of measuring the pore sizes and their locations very accurately. On the downside, this method is extremely time consuming, so that even small amounts of powder (in the order of some mg) might take several hours to yield a high-quality result.

Since XCT is slow and always yields a large three-dimensional data set which needs to be analyzed, it would be desirable to rely on two-dimensional data (i.e., radiographs) which can be acquired in a fraction of the time needed for XCT. The time needed to acquire a high-quality radiograph depends on several parameters, but it is safe to say, that one second is a good estimate for the average exposure time. The image quality requirements for such a radiographic setup have already been formulated by Jaenisch et al.~\cite{Jaenisch2020}. The hardware for a possible radiographic in- or at-line solution already exists, but the result relies at least in equal measure on a proper and fast image processing algorithm.

\citet{wang2020machine} and \citet{meng2020machine} review current applications of machine learning in additive manufacturing; it is seen that variations of CNN-based models (convolutional neural networks) see frequent use for image segmentation tasks in AM. Subsequently, \citet{gobert2020porosity} initiate a pipeline for porosity segmentation of metallic AM specimens based on Otsu thresholding \cite{otsu} and a CNN to segment porosity from XCT images. While similar in approach, our work rather looks into the possibility of using single radiographs to study porosity in the feedstock.  To enable this, deep learning models are compared and combined with classical image processing approaches to achieve a precise and reliable detection of pores in metallic feedstock powders.
The highlights are as follows:
\begin{enumerate}
    \item High pore defect segmentation accuracy on metallic feedstock powders  (F1-score 0.87) can be achieved from single 2D radiographs.
    \item Synthetic radiographs, which assist in model convergence, are efficiently simulated with \textit{aRTist} \cite{artist}.
    \item Modelling X-ray attenuation before the use of a CNN increases test F1 by 11.4\% compared to the baseline CNN.
\end{enumerate} 
These findings are promising for future high throughput porosity analysis of feedstock powders.

\section{Materials and Methods} 
\label{Materials and Methods}
\subsection{X-ray computed tomography data} 
\label{Data}
The openly accessible XCT data set from \textit{Zenodo}~\cite{Schumacher2022} was used for this study. This data set contains an XCT scan of 11 individual metallic powder particles made from an (Mn,Fe)$_2$(P,Si) alloy~\cite{Miao2020} with an average density of {$6.4~g/cm^3$}. The powder was already used and described in detail in~\cite{Jaenisch2020}.
The particle size range in the XCT data set is about 100~--~150~µm. The pore size analysis of the reconstructed 3D data set is based on the Otsu method for automatic image thresholding \cite{otsu} and was achieved with \textit{VG-Studio MAX 3.4.5}~\cite{VG-company}. It yields a large variety of different pore sizes with equivalent pore diameters of up to 75~µm. 
Although the particle size range is slightly larger than commonly used in feedstock powders for AM purposes~\cite{Vock2019}, the mono-layered (conveyor belt-like) arrangement and the mostly spherical particles make this data set ideal for developing a 2D pore segmentation algorithm.

\begin{figure}
\centering
\begin{subfigure}{\linewidth}
  \centering
    \includegraphics[angle=0, width=\linewidth]{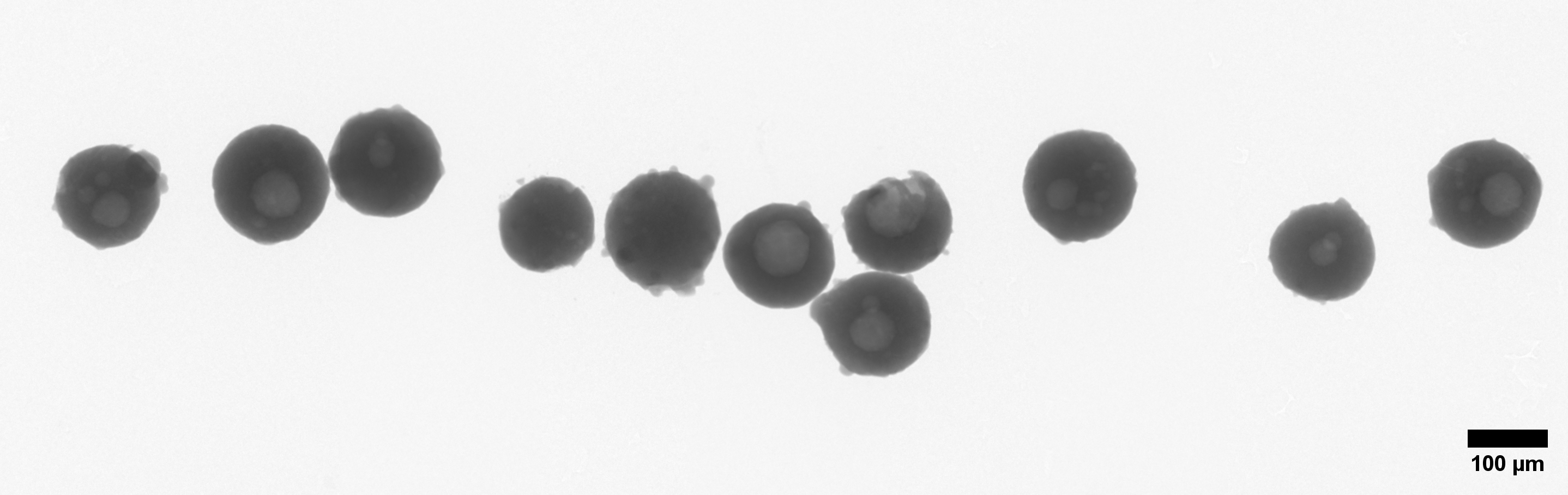}
  \caption{}
  \label{fig:stitch}
\end{subfigure}\vspace{10px}
\begin{subfigure}{\linewidth}
  \centering
   \includegraphics[angle=0, width=\linewidth]{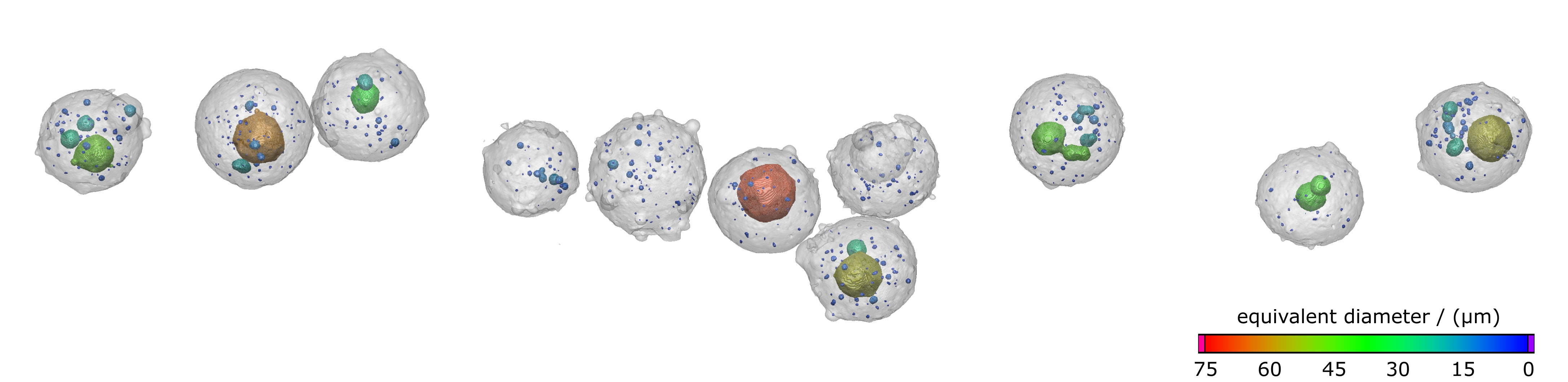}
  \caption{}
  \label{fig:helper}
\end{subfigure}
\caption{A single, stitched 2D projection (\subref{fig:stitch}) of the 801 projections used for 3D reconstruction, and (\subref{fig:helper}) the same view over a 3D rendering of the segmented volume; hue encodes the equivalent diameter of pores. Appendix~\ref{app:render} shows the full-size image.}
\label{fig:gthelper}
\end{figure}

A 2D projection of the analyzed 3D data set (Figure~\ref{fig:helper}) allows to asses the positions and sizes of particles and pores and hence compared it with a corresponding single 2D radiographic projection image (Figure~\ref{fig:stitch}).
An overlay of the radiograph and the analyzed 2D projection was achieved by affine image registration and represents the ground truth data which was used to train, validate and test the developed algorithm.

\subsection{Simulated X-ray computed tomography data} 
\label{Data simulation}
For pretraining the convolutional neural network to be described in section~\ref{sec:MLapproach}, powder particles were simulated with the \textit{aRTist} software \cite{artist}. The radiographic scene within \textit{aRTist} was set-up in a way that represents the real XCT setup (see Sections~\ref{Data} and \ref{Preprocessing of images}) best. 
Particle and pore sizes are drawn from the measured sample of $11$ particles. The pore sizes are given by the porosity analysis of the reconstructed 3D volume~\cite{Schumacher2022} which computes accurate equivalent diameters of the pores, c.f.\ Figure~\ref{fig:dia_hist}. As visualized by the fitted curve on Figure~\ref{fig:dia_hist}, a log-normal distribution captures the real distribution well, 
\begin{align}
    f_X(x) = \frac{1} 
{(x-\theta)\sigma\sqrt{2\pi}}\exp\left[\frac{-(\ln((x-\theta)/\mu))^2}{2\sigma^{2}}\right]
\end{align}
resulting in a mean of $\mu = 3.70$, shape parameter $\sigma= 0.41$ (also the standard deviation of the log of the distribution), and location parameter $\theta= 0.50$. %

Exposure time, X-ray energy, detector settings, and the geometric setup were adjusted to match the ZEISS XCT setup within the limits of \textit{aRTist}.
To drastically increase the amount of data which can be feasibly generated, it is necessary to control \textit{aRTist} efficiently. This is done by establishing a TCP connection with \textit{aRTist} and  controlling its functions and settings remotely --- made possible using an openly available Python library for \textit{aRTist} and manipulating one of the standalone scripts~\cite{artist-lib}. Finally, to urge more robustness regarding particle- and pore shape, the artificial images were warped by a small, smooth, elastic deformation following~\cite{simard}; resulting in images like Figure~\ref{fig:simulated}.

\subsection{Preprocessing, labelling and data split} 
\label{Preprocessing of images}
\begin{figure*}
\centering
\begin{subfigure}{.3\linewidth}
  \centering
  \includegraphics[trim={25 30 30 25}, clip,  width=\linewidth]{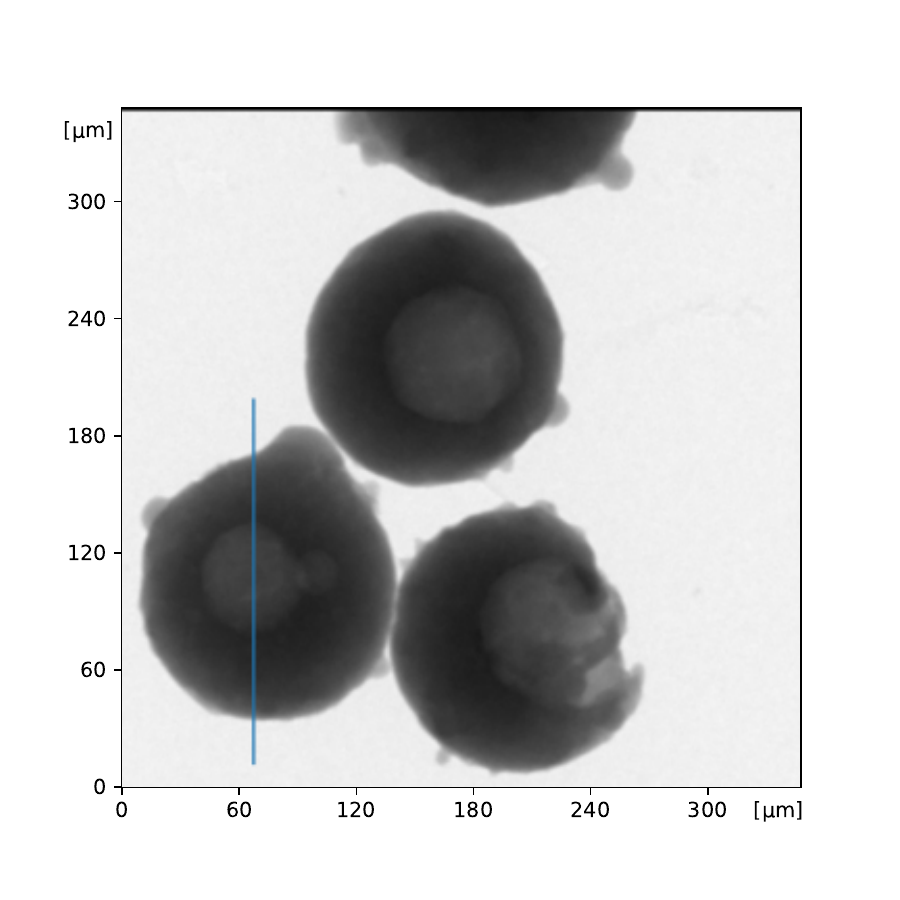} %
  \caption{}
  \label{fig:cutout3}
\end{subfigure}%
\begin{subfigure}{.3\linewidth}
  \centering
  \includegraphics[trim={25 30 30 25}, clip, width=\linewidth]{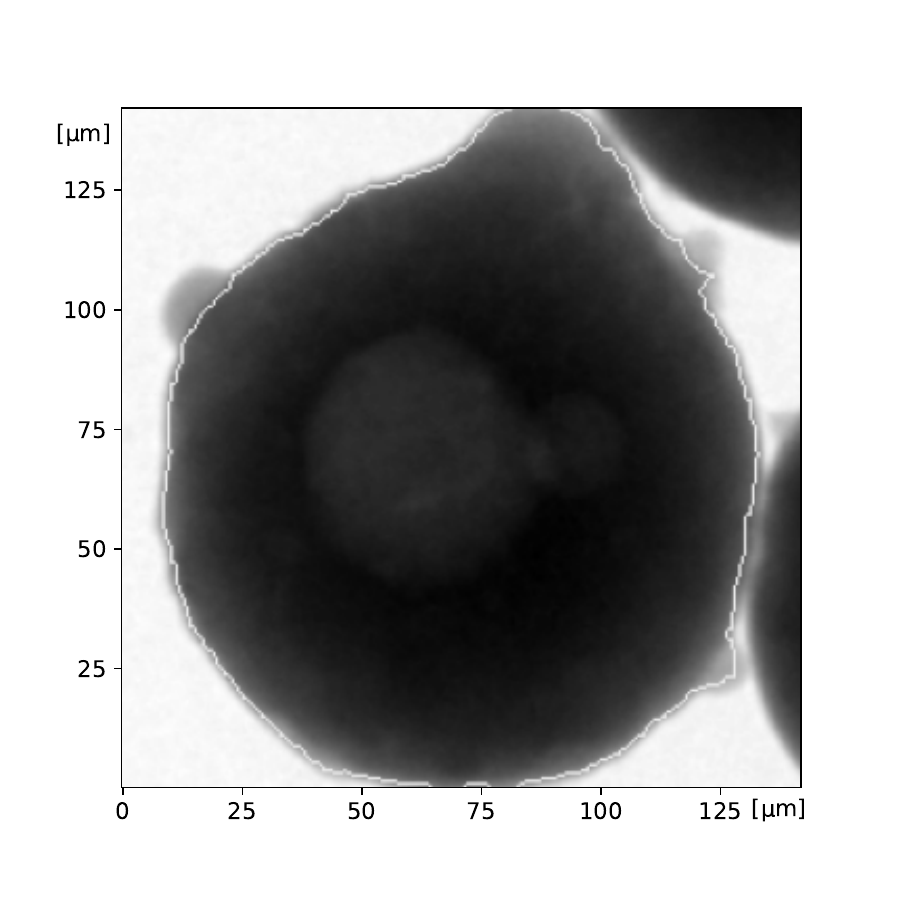}
  \caption{}
  \label{fig:cutout4}
\end{subfigure}\\
  \centering
  \begin{subfigure}{0.32\linewidth}
    \centering
    \hspace{-10px}
    \includegraphics[trim={0 15 30 25}, clip, width=\linewidth]{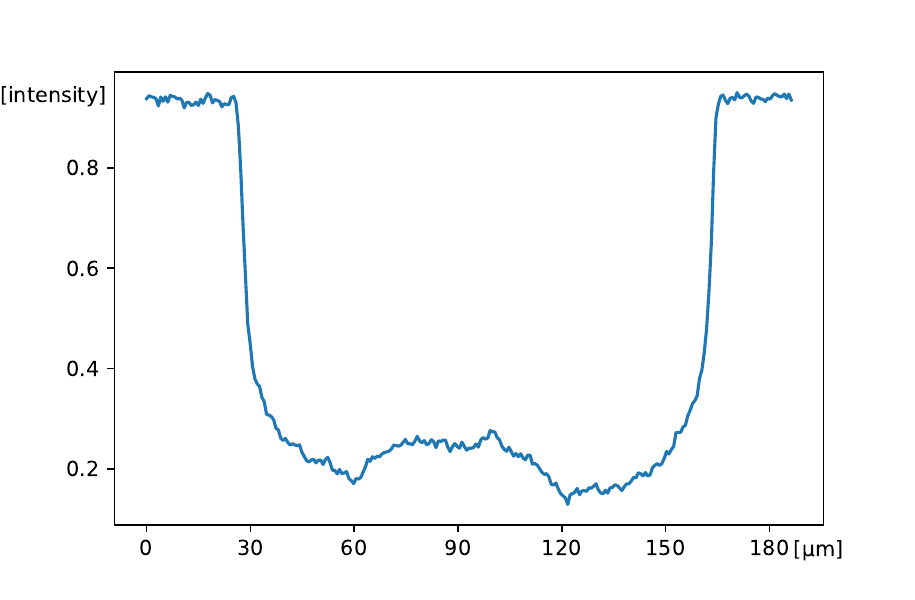}
    \caption{}
    \label{fig:real_attenuation}
  \end{subfigure}
  \begin{subfigure}{0.32\linewidth}
    \centering
        \vspace{10px}
        \hspace{-7px}
    \includegraphics[trim={0 -7 0 0}, clip, width=0.98\linewidth]{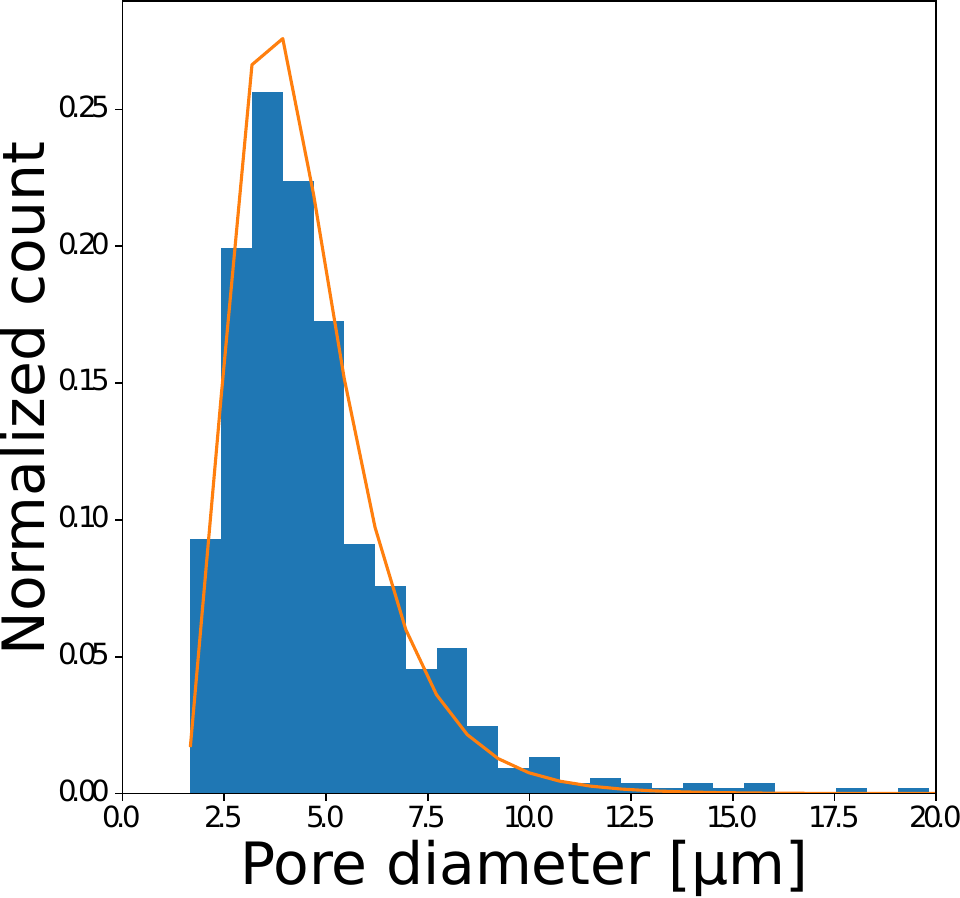}
    \vspace{-7px}
    \caption{}
    \label{fig:dia_hist}
  \end{subfigure}
    \begin{subfigure}{0.3\linewidth}
    \centering
    \includegraphics[trim={25 30 30 25}, clip, width=\linewidth]{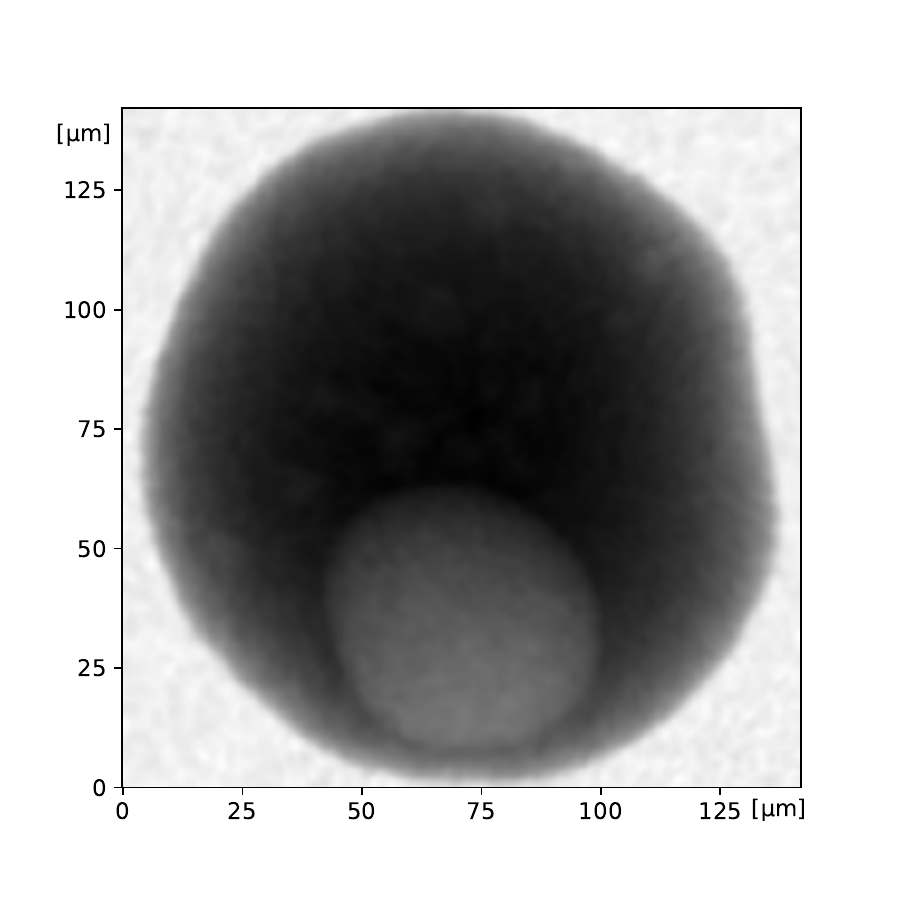}
    \caption{}
    \label{fig:simulated}
  \end{subfigure}
  
\caption{Particles from a single projection. (\subref{fig:cutout3}) The normalized original;  (\subref{fig:cutout4}) the cutout with a faint single particle mask outline in white;
(\subref{fig:real_attenuation}) the intensity profile of a preprocessed particle indicated by the blue line in (\subref{fig:cutout3}); (\subref{fig:dia_hist}) a normalized histogram of the diameter of the particles observed fitted with a log-normal distribution; (\subref{fig:simulated}) a simulated radiograph example.} 
\label{fig:attenuation}
\end{figure*}

The individual images of the stitched radiographic projection image in Figure~\ref{fig:stitch} are standardized on a per-image basis by linearly scaling to mean $0$ and variance $1$, and subsequently normalized to a 0--1 range, in order to account for slight intensity and noise variations caused by e.g. fluctuations of target current.
Further, image cutouts are made as shown in Figure~\ref{fig:cutout4} for a more balanced class ratio %
and reduced memory requirements.
The cutout algorithm is based on labeling connected components of particle masks followed by computing and resizing the most fitting square non-intruding bounding box to a 256x256 resolution. Image size choice is made from considering particle extents. %
Masks isolating single particles are created by combining a Canny edge image with masks from thresholding; the resulting mask is faintly outlined on Figure~\ref{fig:cutout4}. 
Figure~\ref{fig:real_attenuation} shows an example of the intensity profile of Figure~\ref{fig:cutout3} where the present pore leads to a convex profile shape in the center of the particle.

In order to train, validate and test the developed models, the available data is separated into 3 subsets: 55\% for training, 18\% for validation, and 27\% for testing purposes; respectively 6, 2, and 3 particles.

\subsection{Highlighting pores by distance to the boundary} 
\label{sec:sdf}
The pores appear in the cutouts as local bright areas, as seen in Figure~\ref{fig:cutout4}. To highlight these areas, we propose to modify the particle's intensity in each point as a function of the shortest distance to the boundary of the particle. This concept, which effectively flattens the background signal of the particle itself, will be explained in the following.

\begin{figure*}
\centering
\begin{subfigure}{.3\linewidth}
  \centering
  \includegraphics[trim={25 30 30 25}, clip, width=1\linewidth]{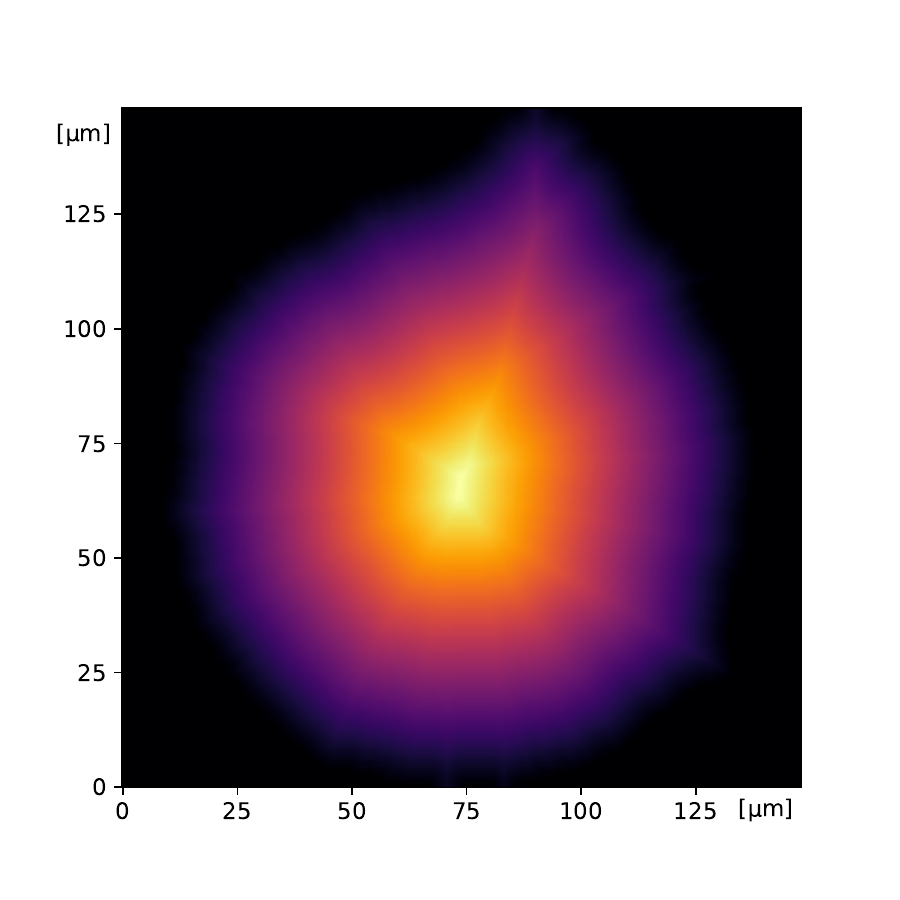}
  \caption{}
  \label{fig:edt}
\end{subfigure}%
\begin{subfigure}{.3\linewidth}
  \centering
  \vspace{3px}
  \includegraphics[trim={0 0 0 -38}, clip, width=1.08\linewidth]{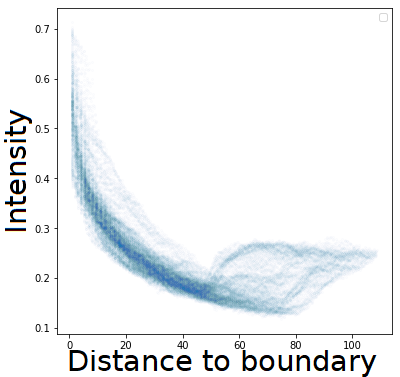}
  \caption{}
  \label{fig:dist_original}
\end{subfigure}\\%
\begin{subfigure}{.3\linewidth}
  \centering
  \includegraphics[trim={25 30 20 25}, clip, width=1.012\linewidth]{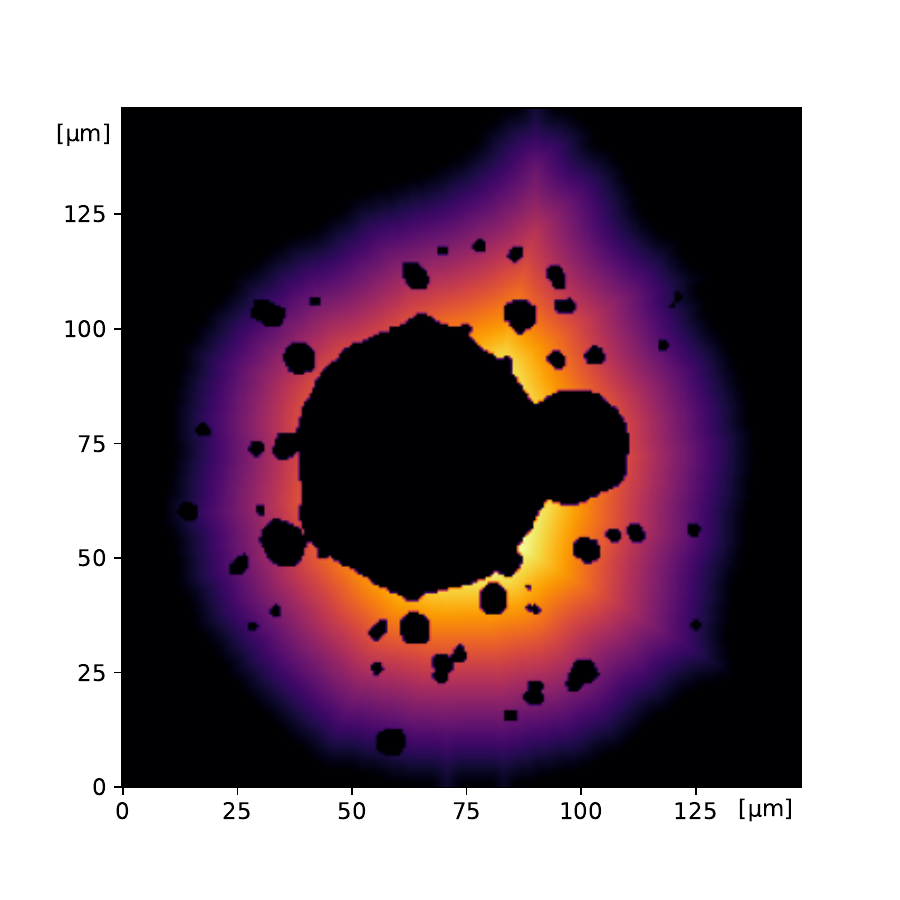}
  \caption{}
  \label{fig:edt2}
\end{subfigure}
\begin{subfigure}{.3\linewidth}
    \centering
        \vspace{-1px}
        \hspace{-14px} %
\includegraphics[trim={0 0 0 -47}, clip, width=1.05\linewidth]{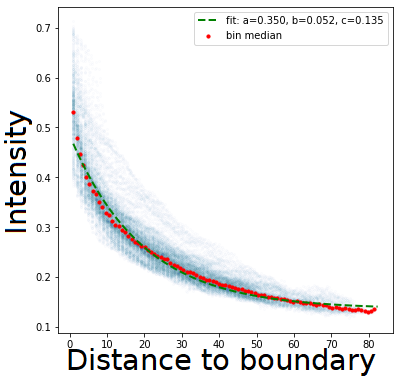}
    \vspace{-4px}\caption{}
    \label{fig:sdfmedian}
\end{subfigure}
\begin{subfigure}{.3\linewidth}
      \centering
      \includegraphics[trim={25 30 30 25}, clip, width=\linewidth]{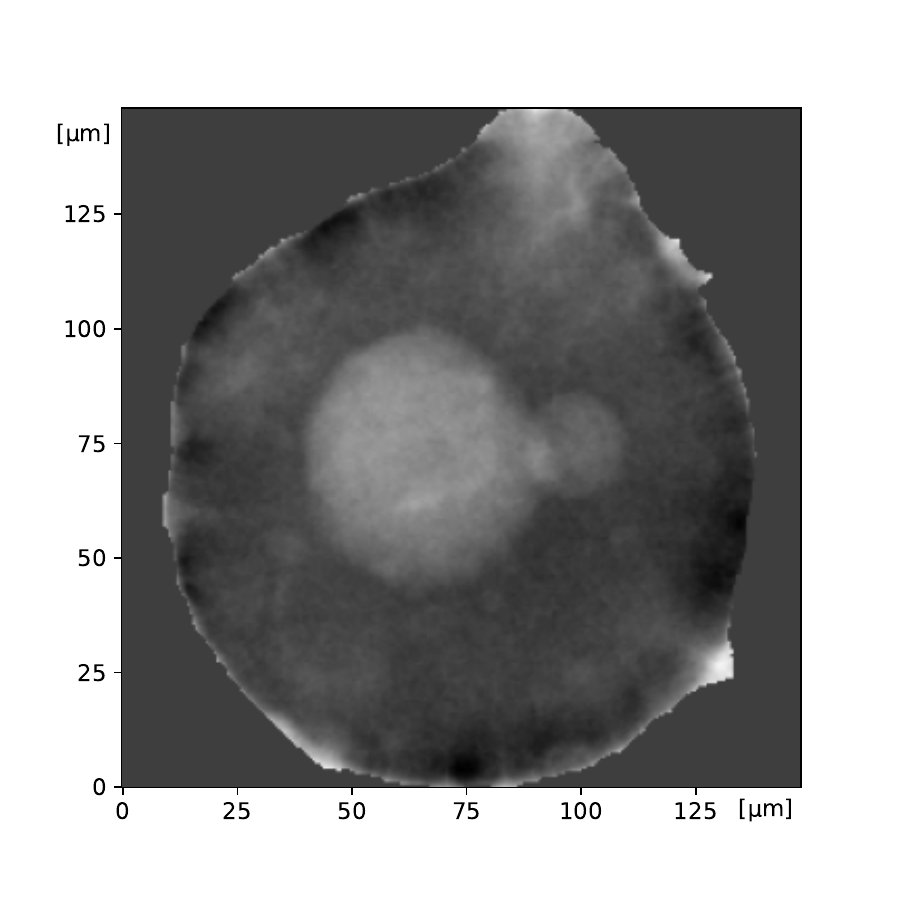}
  \caption{}
  \label{fig:cutoutsubtracted}
\end{subfigure}%
\caption{Cutout subtraction and pores for the particle shown in Figure~\ref{fig:cutout4}. (\subref{fig:edt}) The signed distance field; (\subref{fig:dist_original}) a scatter plot of observed intensities vs distance to boundary for distances greater than 0; (\subref{fig:edt2}) the distance field overlaid with the ground-truth pores; (\subref{fig:sdfmedian}) similar scatter but without assumed pore pixels, with aggregated points in red and a fit on these to Eq.~\eqref{eq:fit} in green; (\subref{fig:cutoutsubtracted}) the image of the particle with the resulting attenuation model subtracted.}
\label{fig:2edt}
\end{figure*}

The intensity profile visualized in Figure~\ref{fig:real_attenuation} shows how intensity decays as X-rays propagate through a gradually thicker medium, i.e. the particle, as it largely follows Lambert-Beer's law. The high contrast between particle and background intensities allows to easily create particle masks by simple thresholding. Conceptually, if we were able to remove the intensity influence of a pore-free particle, which always has a concave profile shape similar to Figure~\ref{fig:real_attenuation}, simple intensity thresholding could be used for the segmentation of the pore as well. Mainly due to applying a linear transformation during preprocessing, we are interested in the more general form,
\begin{align}
I(x)= a e^{-bx}+c,
\label{eq:fit}
\end{align}
To learn the parameters $a$, $b$, and $c$, the signed distance field (SDF) on each particle mask (mapping every pixel to the shortest distance to the boundary) is computed based on SciPy's \verb|ndimage.morphology.distance_transform_edt|. With each pixel having both an intensity and a corresponding distance measure, the attenuation function (Eq.~\eqref{eq:fit}) can be learned. However, to increase robustness of the fit, it is performed on 95 bins placed uniformly on the x-axis in the range $[1;\max (\text{distance})]$. The corresponding intensity value of a bin is computed as the 40th percentile of the aggregated intensity values. These hyperparameters (bin count and percentile) are estimated empirically; note (from Figure~\ref{fig:sdfmedian}) that a percentile slightly lower than the median helps to compensate for some boundary shape deformation. In Figure~\ref{fig:cutoutsubtracted}, the result of subtracting this final distance map (with fit values given in Figure~\ref{fig:sdfmedian}) is shown.

\subsection{Models for pore segmentation}\label{sec:Model}
This paper presents four different models which all seek to segment the pores within a particle from a single 2D view. The flow chart in Appendix~\ref{fig:overview} shows the process pipeline leading to the four proposed models. The models differ in complexity, approach (probabilistic or deterministic), accuracy, and finally in efficiency. The proposed deterministic models are the \textsc{Local threshold} and \textsc{Attenuation-Adjusted threshold} models (described in detail in sections \ref{sec:local} and \ref{sec:global}, respectively). The former simply applies a Gaussian-based local threshold. The latter computes an "ideal" (i.e., pore-free) particle, subtracts this from the 2D radiograph, and leaves pore pixels with a relatively large intensity. This "ideal" particle (and hence the whole model) is iteratively refined until the pore predictions converge and model accuracy is optimal. 

Both of the probabilistic models are based on a variation of the UNet neural network architecture; they solely differ based on their input. The simpler model is denoted \textsc{UNet} model (Section \ref{sec:unet}) and uses just the pre-processed 2D radiography data as input. The \textsc{Combined} model (Section \ref{sec:MLapproach}) applies the UNet architecture to an intermediate result of the last iteration of the \textsc{Attenuation-Adjusted threshold} model using the "ideal" particle from Section \ref{sec:global}.

In the following, we denote counts of True/False Positive (TP/FP) as respectively correctly and incorrectly segmented pore pixels, while True/False Negative (TN/FN) denote correctly/incorrectly segmented non-pore pixels. Then, the $\text{F1-score}=\rfrac{2\text{TP}}{(2\text{TP}+\text{FP}+\text{FN})}$ is utilized as an accuracy measure. 

\subsubsection{The local threshold model}\label{sec:local}
\begin{figure*}
\centering
\begin{subfigure}{.33\linewidth}
  \centering
  \includegraphics[trim={25 30 30 25}, clip, width=\linewidth]{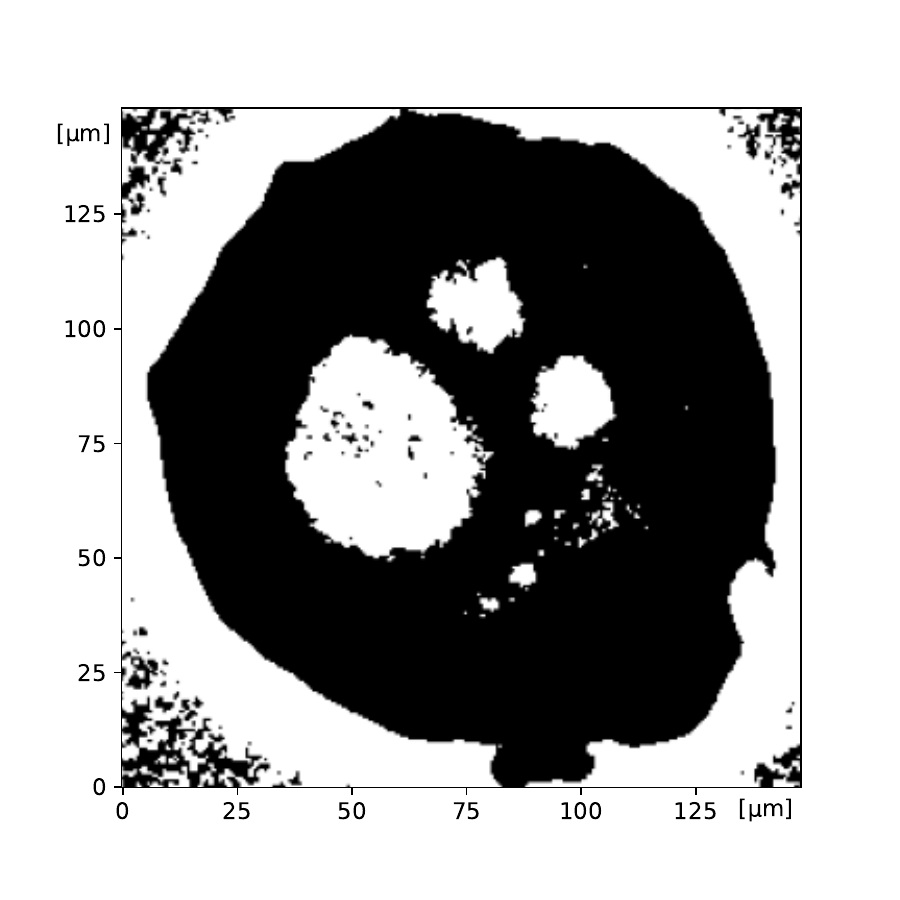}
  \caption{}
  \label{fig:local_thr}
\end{subfigure}%
\begin{subfigure}{.33\linewidth}
  \centering
  \includegraphics[trim={25 30 30 25}, clip, width=\linewidth]{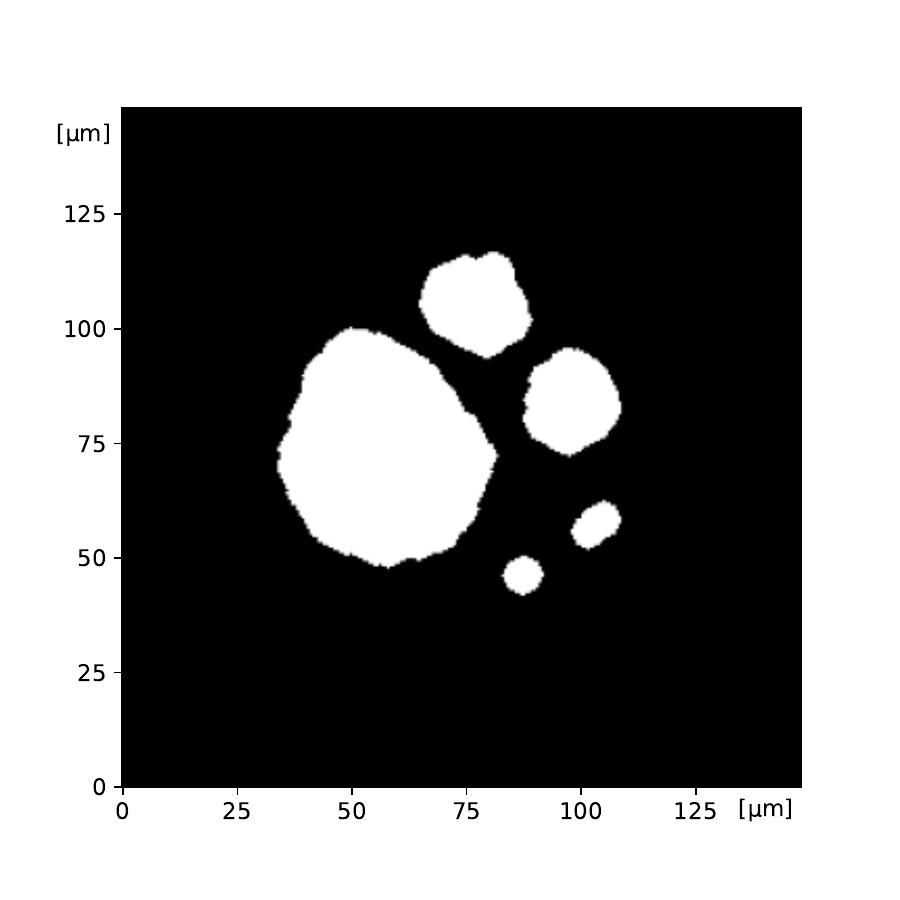}
  \caption{}
  \label{fig:local_thr_post}
\end{subfigure}%
\begin{subfigure}{.33\linewidth}
    \centering
    \vspace{-10px}
    \includegraphics[width=1.1465\linewidth]{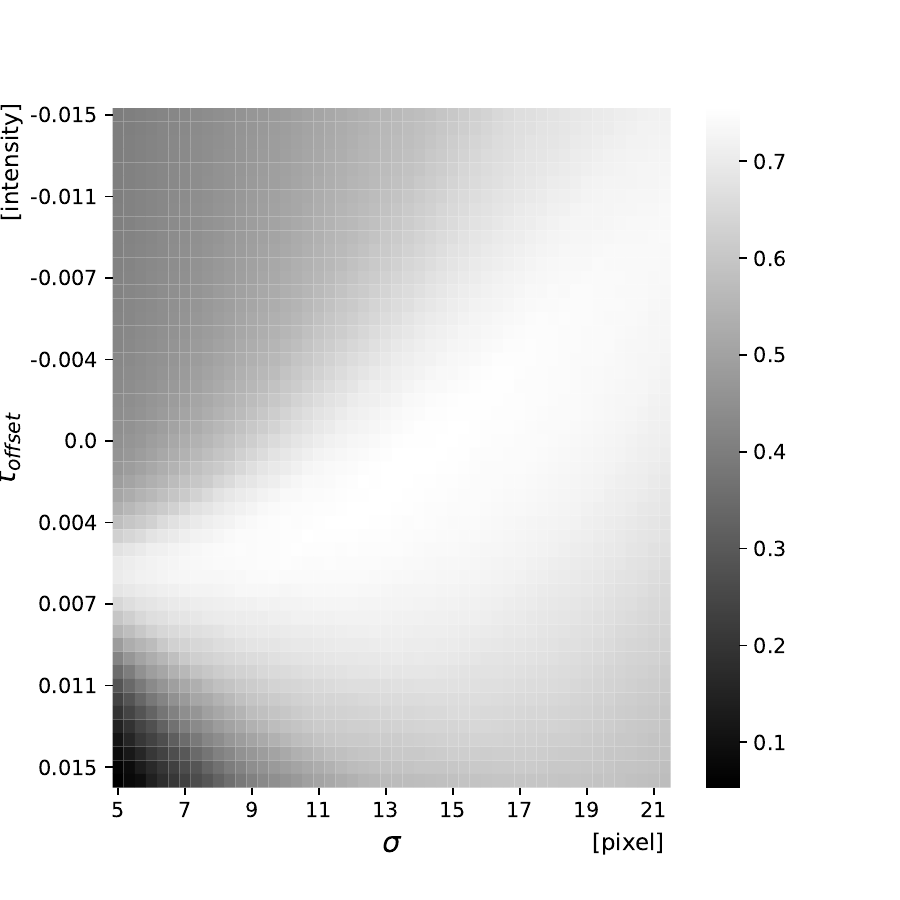} %
    \vspace{-25px}
    \caption{}
    \label{fig:gridsearch}
\end{subfigure}%
    \caption{
    (\subref{fig:local_thr}) The result of local thresholding; (\subref{fig:local_thr_post}) the denoised image; (\subref{fig:gridsearch}) a gridsearch over the validation set showing F1-score plotted versus hyperparameter values.}
\label{fig:local_thr_gridsearch}
\end{figure*}

The first model is an adaptive thresholding method defined as follows: 
\begin{align}
&    I(\vec x) > \left(I*G_{\sigma}\right)(\vec x) + t_{\text{offset}}\\
& \text{with }\,G_{\sigma}(\vec x) = \frac{1}{\sigma\sqrt{2\pi}}\exp\left(\frac{|\vec x|^2}{2\sigma^2}\right)
\end{align}
The resulting boolean images are very noisy. Postprocessing is done using mathematical morphology --- here a short, fast sequence with eroding and dilating operations. 
A gridsearch for optimal values of $\sigma$ and $t_{\text{offset}}$ is shown on Figure~\ref{fig:gridsearch}.

\subsubsection{The attenuation-adjusted threshold model} \label{sec:global}
Using the attenuation-inspired subtraction method described in \ref{sec:sdf}, an ideal pore-free particle is modelled. As a consequence, the resulting images are particularly bright in the region of pores --- which may be segmented via thresholding. The choice of threshold value is a balance act done by maximizing F1 on the validation set.
Since the attenuation model is less precise near the uneven boundary, it is difficult to set a threshold to both, include the correct pores while excluding boundary points. However, we observe the large majority of pores not being close to the boundary, so indications near the boundary may be ignored --- this is done by removing regions near the boundary calculated as a dilation of the particle-boundary curve. Finally, the attenuation-subtracted images also contain noise, which is reduced by a sequence of \verb|scikit-image| morphology operations containing binary erosion, dilation, and \verb$remove_small_holes$. Lastly, "pore-like" priors are enforced (placing a lower bound of region area relative to the bounding box, and forcing that the centroid has the same region label) based on \verb$skimage.measure.regionprops$. 

The location of the pores influences the estimation of the attenuation model, e.g. in Figure~\ref{fig:edt2} the distance field is shown in relation to the ground-truth locations of the pores. To reduce the effect of the pores, we have used an iterative algorithm, where we iteratively estimate the attenuation model, use the threshold method described above to estimate the pores, and then re-estimate the attenuation model ignoring intensity values overlapping with pore-pixel candidates. We iterate until convergence, which in our experience often requires 4--6 steps. The effect on the initial model shown in Figure~\ref{fig:dist_original} is shown in Figure~\ref{fig:sdfmedian}. The final model we call the \textsc{Attenuation-Adjusted threshold} model.

\subsubsection{The UNet model}
\label{sec:unet}
A variant of the UNet architecture~\cite{unet} is applied, including batch normalization before each block of convolutional layers. The network was pretrained on abundant artificial data described in Section~\ref{Data simulation}. The artificial data and the preprocessed radiography data (see Section~\ref{Preprocessing of images}) were augmented through horizontal and vertical flips.  Data augmentation, image normalization and making tight cutouts were crucial to accommodate the data for the neural network. Our experience further shows that pretraining is useful for convergence in cases with limited ground-truth data, such as the one reported in this article, and for decreasing the following training time.
The training procedure is based on the PyTorch automatic differentiation package (autograd) by minimizing loss subject to the Adamax optimizer~\cite{kingma2017adam}. The choices of UNet depth, the optimizer's parameters, the loss function, and estimation of the model's generalization error are done using the validation data. This set also enables early stopping of training at the minimum validation loss, giving a regularizing effect and avoids overfitting to the limited training data. The optimal loss function was found to be binary cross-entropy, with the optimal network being a 54-layer deep UNet~\cite[Figure~44]{andreas} with 4 contracting blocks, and Adamax-parameters: \texttt{lr=1e-4, betas=(0.9, 0.999), eps=1e-08, weight\_decay=0}. The loss history is shown on Figure~\ref{fig:pretrained_real_history_raw}.

\subsubsection{The combined model}
\label{sec:MLapproach}
\begin{figure}
\centering
\begin{subfigure}{.65\linewidth}
  \centering
\includegraphics[trim={0 0 -15 0}, clip, width=\linewidth]{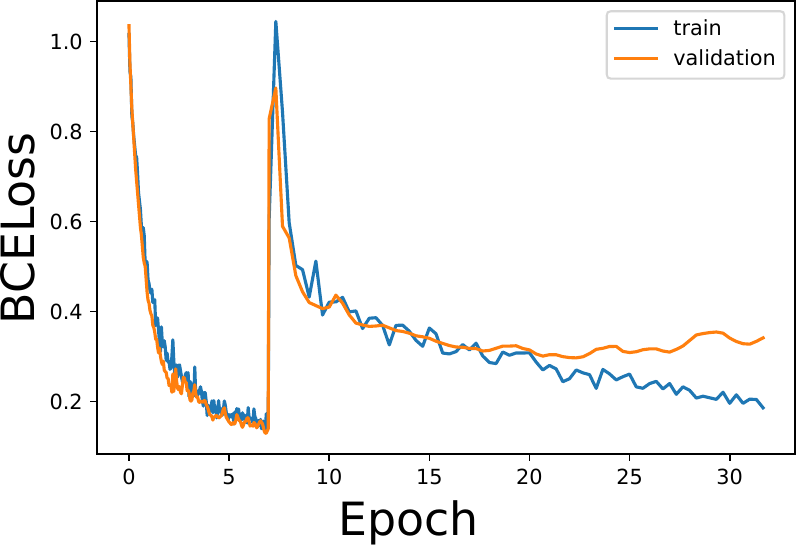}\vspace{-5px}
  \caption{}
  \label{fig:pretrained_real_history_raw}
\end{subfigure}\vspace{10px}
\begin{subfigure}{.65\linewidth}
  \centering
\includegraphics[trim={0 0 -15 0}, clip, width=\linewidth]{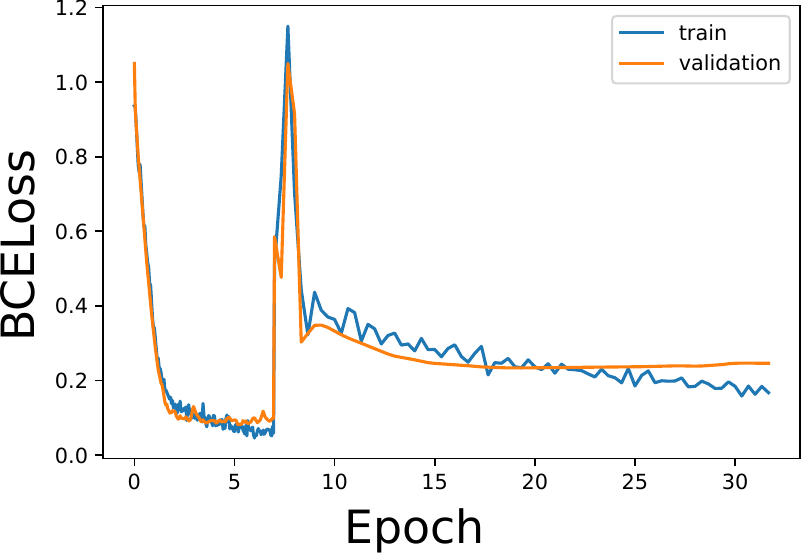}\vspace{-5px}
  \caption{}
  \label{fig:pretrained_real_history_sdf}
\end{subfigure}
\caption{Evolution of training and validation loss for (\subref{fig:pretrained_real_history_raw}) UNet, and (\subref{fig:pretrained_real_history_sdf}) combined model. Training is done with first 7 epochs of artificial \textit{aRTist} data, and then 25 epochs of real data. After augmentation, the training set sizes are 200 and 24 images, respectively. The jump in loss value is due to swapping to real data --- and despite the large peak, pretraining was found useful for convergence and for decreasing the training time. 
}
\label{fig:training_history}
\end{figure}

Our final proposed model is a UNet trained and applied on attenuation-subtracted images. Here, the same network architecture and training procedure as described in Section~\ref{sec:unet} was used. See Figure~\ref{fig:cutoutsubtracted} for an example input image; these are by-products from the \textsc{Attenuation-Adjusted Threshold} model. As before, Figure~\ref{fig:pretrained_real_history_sdf} shows the loss history. 

\section{Results and discussions}
To evaluate the quality of the segmentation models, the confusion matrix was calculated on test data which was initially separated from the data set and thus has not taken part in fitting, training, or validating the models. %
To highlight various aspects of the quality of the segmentation, different combinations of TP/FP/TN/FN counts can be used; we will use the 
true positive rate $\text{TPR}=\rfrac{\text{TP}}{\text{TP}+\text{FN}}$, 
false negative rate $\text{FNR}=\rfrac{\text{FN}}{\text{FN}+\text{TP}}$,
true negative rate $\text{TNR}=\rfrac{\text{TN}}{\text{TN}+\text{FP}}$,
false positive rate $\text{FPR}=\rfrac{\text{FP}}{\text{FP}+\text{TN}}$,
and the previously mentioned  $\text{F1-score}=\rfrac{2\text{TP}}{(2\text{TP}+\text{FP}+\text{FN})}$.
Table~\ref{tab:iterationF1} shows the effect of the iterative update of the attenuation-Adjusted threshold model --- here, F1-score is seen to increase monotonically with little improvement after 4 iterations.  
\begin{table}[width=1\linewidth,cols=7]
    \caption{Evolution of test set F1-score over several iterations during the \textsc{Attenutation-Adjusted threshold} model.}
    \label{tab:iterationF1}
    \centering
    \begin{tabular*}{\tblwidth}{r|c|c|c|c|c|c}
    Iteration & 1 & 2 & 3 & 4 & 5 & 6 \\ \hline
    F1-score & $0.720 $ & $0.791 $ & $0.819 $ & $0.826 $ & $0.827 $ & $0.827 $
\end{tabular*}
\end{table}
The resulting fit provides estimation of the $a$, $b$, and $c$ parameters of Eq.~\eqref{eq:fit} for a particle without pores, where $a$ and $c$ are related to the affine transformation of the intensities performed both by the imaging hardware and our preprocessing, and $b$ is related to the attenuation coefficient of the particle material. The basis of Eq.~\eqref{eq:fit} is the distance field which is a model of the intersection length between the particle and the X-ray. The model assumes that the particles are spherical, and while certainly not all are, as seen in Figures~\ref{fig:stitch} and~\ref{fig:cutout3}, many are. Nevertheless, its validity is uncertain, and hence, further investigation is needed to ascertain the usefulness of $b$ as a material parameter in this context. 

\begin{figure*}
\centering
\begin{subfigure}{.24\linewidth}
  \centering
  \includegraphics[trim=20 40 20 20,clip,width=\linewidth]{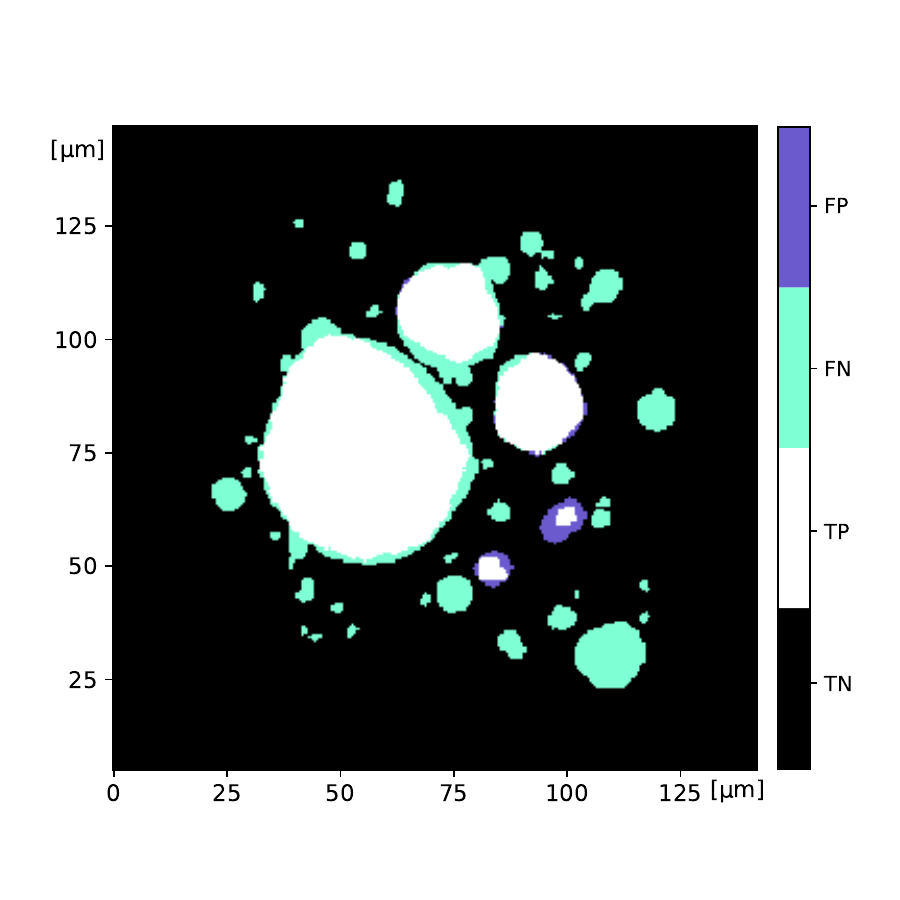}%
  \caption{}
  \label{fig:error_types_loc}
\end{subfigure}
\begin{subfigure}{.24\linewidth}
  \centering
  \includegraphics[trim=20 40 20 20,clip,width=\linewidth]{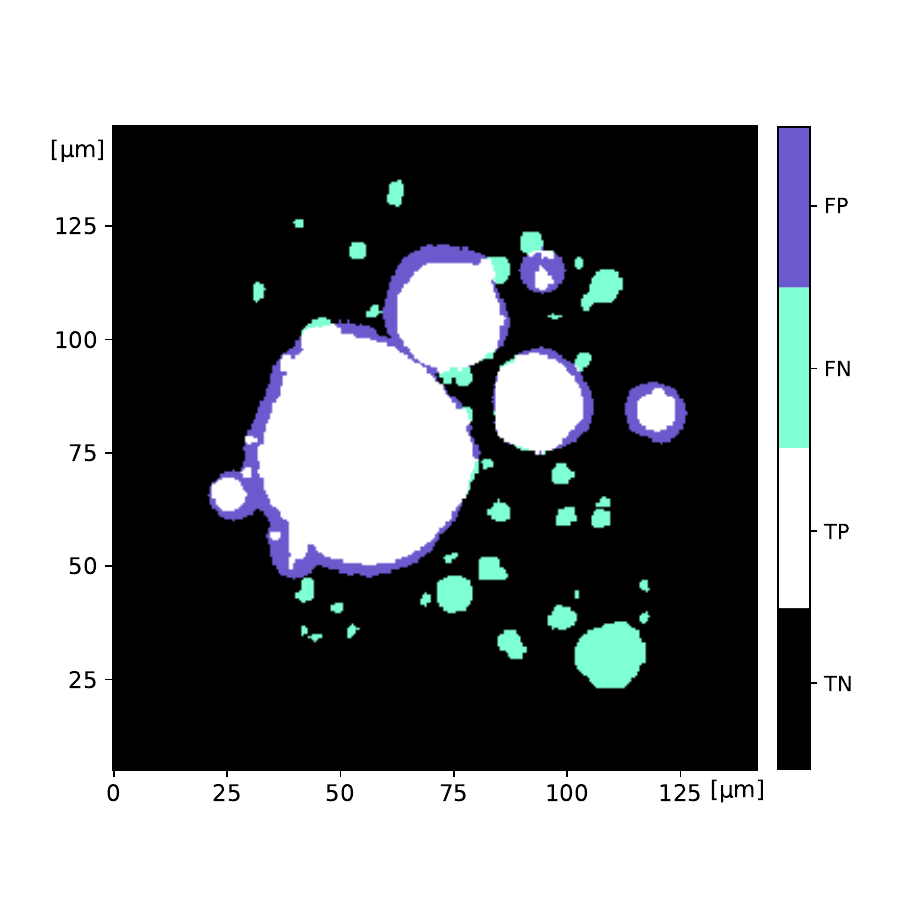}
  \caption{}
  \label{fig:error_types_thresholding}
\end{subfigure}
\begin{subfigure}{.24\linewidth}
  \centering
  \includegraphics[trim=20 40 20 20,clip,width=\linewidth]{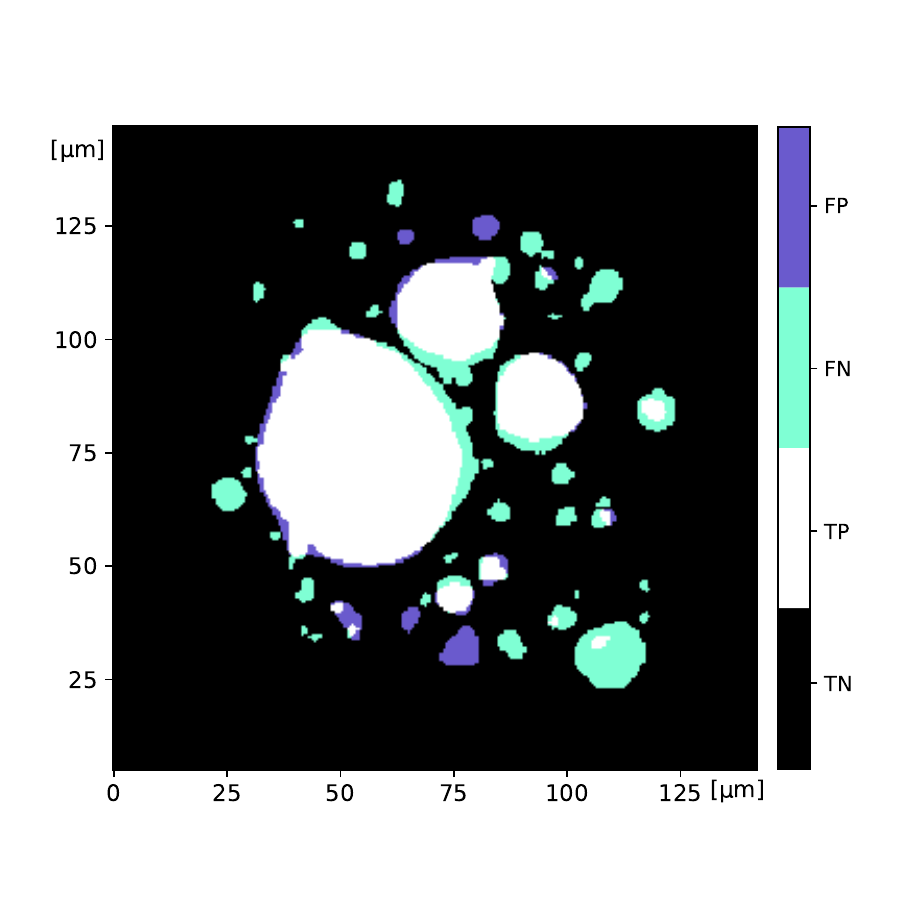}%
  \caption{}
  \label{fig:error_types_unet}
\end{subfigure}
\begin{subfigure}{.24\linewidth}
  \centering
  \includegraphics[trim=20 40 20 20,clip,width=\linewidth]{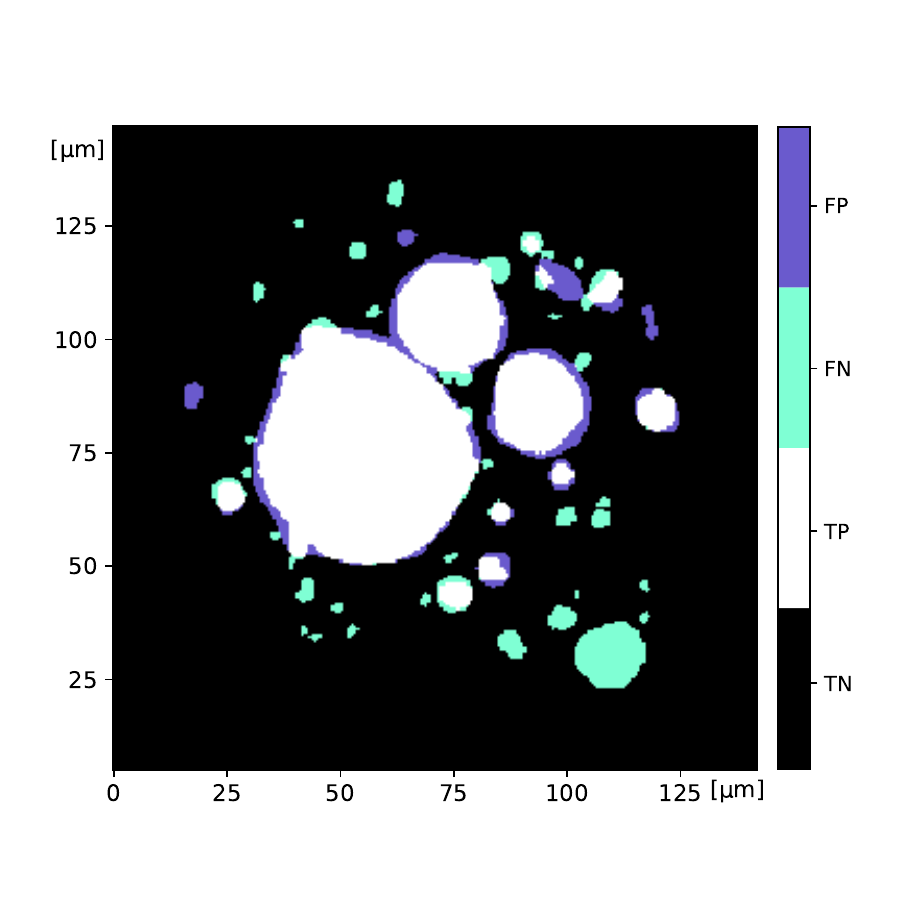}
  \caption{}
  \label{fig:error_types_combined}
\end{subfigure}
\caption{Error types for segmentation on test data. %
While dangerous to generalize from a single example, these images provide a qualitative aspect to evaluating the models. (\subref{fig:error_types_loc}) Local threshold; (\subref{fig:error_types_thresholding}) attenuation-adjusted threshold; (\subref{fig:error_types_unet}) UNet; (\subref{fig:error_types_combined}) combined model. }
\label{fig:error_real}
\end{figure*}

A qualitative comparison of the proposed models is given in Figure~\ref{fig:error_real}, where outputs for an example test image are shown --- with pixels colored with respect to the true labels. 

The example indicates that all methods struggle with small pores, that the deterministic models tend to oversegment pores, and that the UNet-based models are not doing well at the boundary of the particle. For the complete test set, a summary of the previously described quantitative measures are listed in Table~\ref{tab:confusionMatrix}.

\begin{table*}[width=1.35\linewidth,cols=7]%
    \caption{Quality assessment of the pore segmentation on unseen test set. $\hat{t}_\mu$ denotes the mean computation time for a single particle when making pore predictions on a batch with $n=64$ particles.}
    \label{tab:confusionMatrix}
    \centering
    \begin{tabular*}{\tblwidth}{r|c|c|c|c|c|c}
    Method & F1 & TNR & FPR & FNR & TPR & $\hat{t}_\mu$ \\
    \hline
    \textsc{Local Threshold} 
    & $0.765$ & $0.978$ &  $0.022$ & $0.289$ & $0.711$ & $0.024$ \\ 
    \textsc{Att.-Adjusted Threshold}
    & $0.826$ & $0.978$ &  $0.022$ & $0.192$ & $0.808$ & $0.324$\\ 
    \textsc{UNet}
    & $0.783$ & $0.944$ &  $0.056$ & $0.160$ & $0.840$ & $0.014$\\ 
    \textsc{Combined}
    & $0.872$ & $0.975$ &  $0.025$ & $0.151$ & $0.849$ & $0.291$
    \end{tabular*}
\end{table*}
Table \ref{tab:confusionMatrix} confirms the observations from Figure~\ref{fig:error_real}; the UNet on the raw cutouts performs worse than the attenuation-adjusted model, but better than the local threshold, and all three are outperformed by the Combined model. 

\begin{figure}
    \centering
    \includegraphics[width=0.65\linewidth]{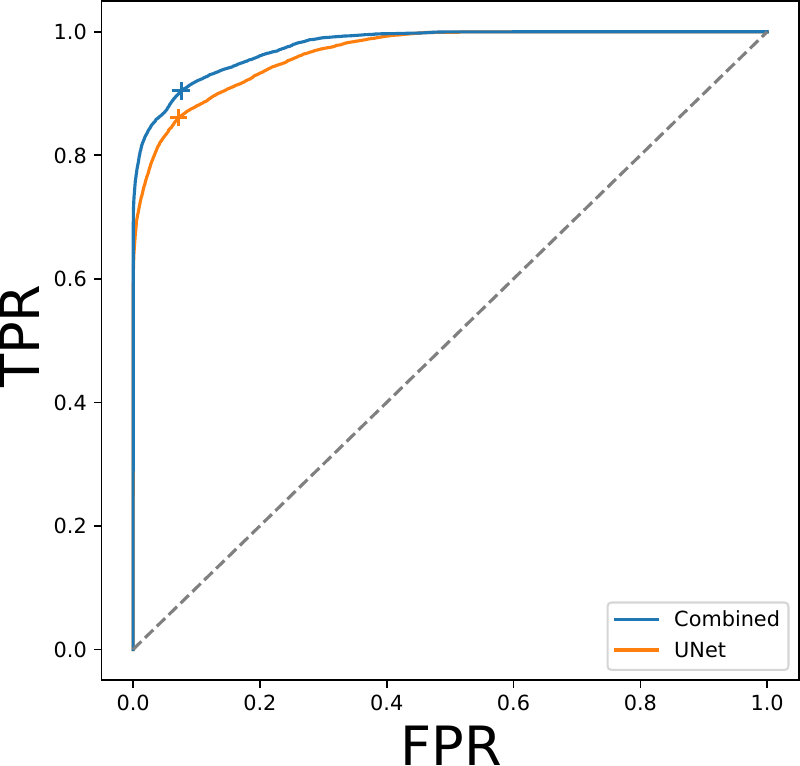}
    \caption{ROC curves for the UNet-based models showing diagnostic ability as the final threshold is varied; markers denote F1-optimized models. As the Local threshold and Attenuation-adjusted threshold methods rely on binary denoising and do not output probabilistic predictions, they are not visualized here.}
    \label{fig:roc_curve}
\end{figure}

Various threshold values $\text{thr}\in[0;1]$ could be applied over the CNN probability outputs. The table is listed with $\text{thr}=0.5$, minimizing training loss which normalizes label influence according to their relative counts (thus, in this case, attributing greater weight to positive examples). Alternatively, one could optimize for F1-score over the validation set, leading to thresholds $\text{thr}_\text{opt-F1} \approx 0.8$ for both models. Such thresholds will naturally provide a greater F1-score, but at the expense of TPR. Hence, the UNet-based models provide more expressive outputs compared to the binary nature of the deterministic ones. Note that for AM feedstock powders, we are mostly interested in the presence of pores and hence \textit{do} wish to put emphasis on positive examples (pore pixels). 
Figure~\ref{fig:roc_curve} shows the corresponding Receiver Operating Characteristic (ROC) curves for the UNet-based models, with measures of area under the curve (AUC-ROC):  \textsc{Combined}: $0.977$; \textsc{UNet} $0.965$. The AUC-ROC measures and the curve appearances give confidence in the models' ability to distinguish between pore/non-pore pixels.  

\begin{figure*}
\centering
\begin{subfigure}{0.325\linewidth}
  \centering
  \includegraphics[trim={0 0 0 0}, clip,  width=\linewidth]{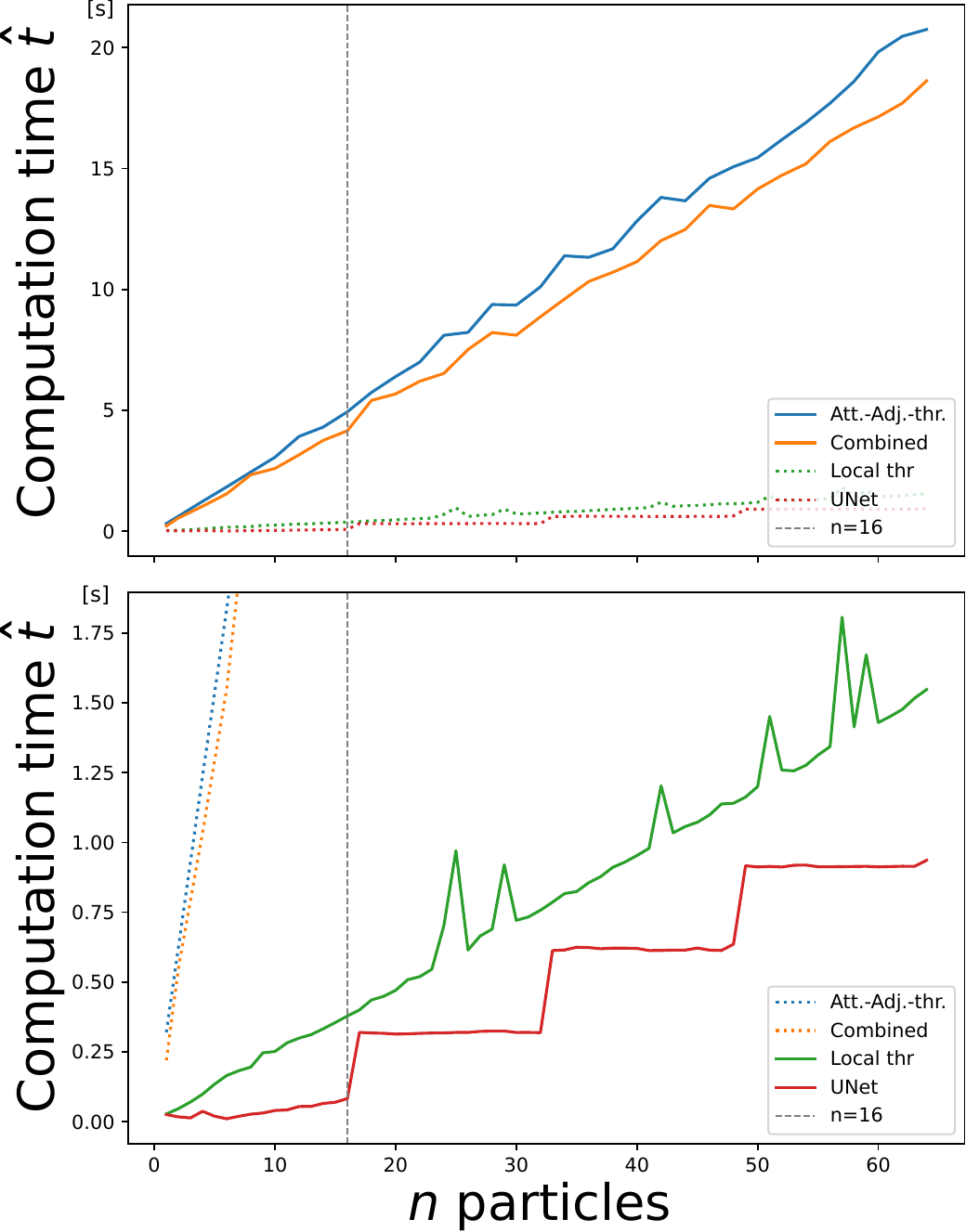}
\caption{}
\label{fig:timings}
\end{subfigure}%
\begin{subfigure}{0.67\linewidth}
  \includegraphics[trim={20, 20, 30, 40}, clip, width=\linewidth]{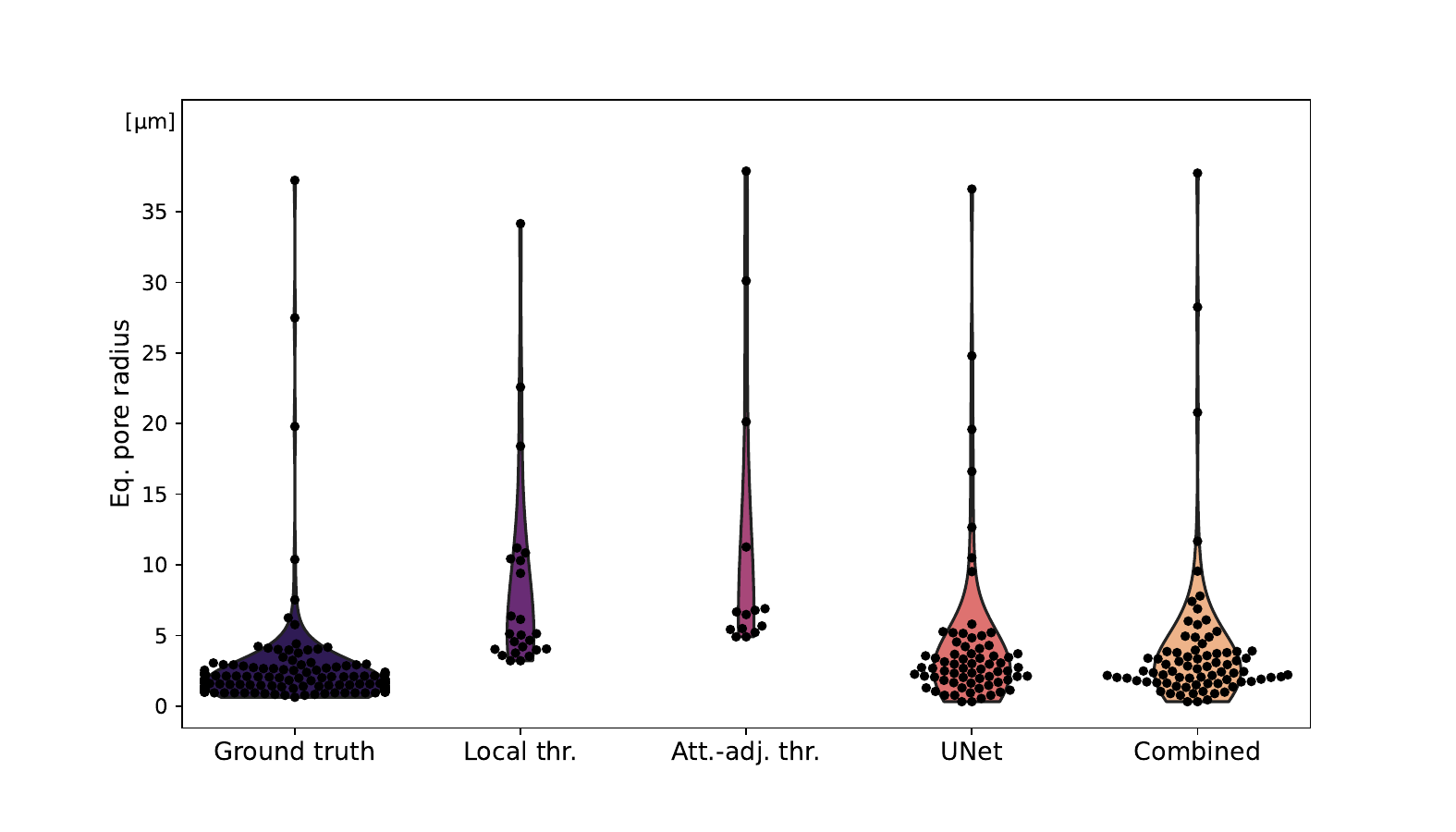}
\caption{} 
\label{fig:areadistreal}
\end{subfigure}%
    \caption{\newline (\subref{fig:timings}) Python computation times for a single run of each model evaluated on $n\in[1;64]$ particles --- split into two plots with varying scale of y-axis for better visibility (scales differ by approximately a factor 10).  \newline (\subref{fig:areadistreal}) Violin plots of segmented equivalent pore radius over the test set. The distribution for each set is symmetrically shown along the vertical axis, and each point denotes a detected pore which has been scattered horizontally to reduce clutter.}
\label{fig:timings_violin}
\end{figure*}

Table~\ref{tab:confusionMatrix} further lists an empirical estimate for the computation time of evaluating the model on a single particle, while Figure~\ref{fig:timings} gives a rough idea of the models' efficiencies as the batch size increases; note: however that this is very setup-dependent. The experiment setup is a GNU/Linux machine running a single-core Intel(R) Xeon(R) CPU @ 2.20GHz, a 16GB Tesla P100, and 16GB RAM. For the UNet based approaches, this allows for a batch size of $n=16$ particles in CUDA memory, leading to the step-like nature of the plot. The most significant contribution to computation time for the two threshold-based approaches is the denoising morphology sequence --- respectively accounting for 58\% and 65\% of the total computation times. As an example pipeline, an initial fast pore approximation (e.g., by \textsc{Local threshold}) can serve as the base for the generation of attenuation-subtracted images which can subsequently be fed into the UNet --- however, it remains needs further investigation whether the local threshold output is suitable for this and can provide similar accuracy. 
While these timings could easily be improved upon --- especially by altering the denoising procedures or, e.g., writing an efficient equivalent implementation in another language  --- they provide a realistic relative ordering of the methods.

\begin{figure*}
    \centering
\includegraphics[width=\linewidth]{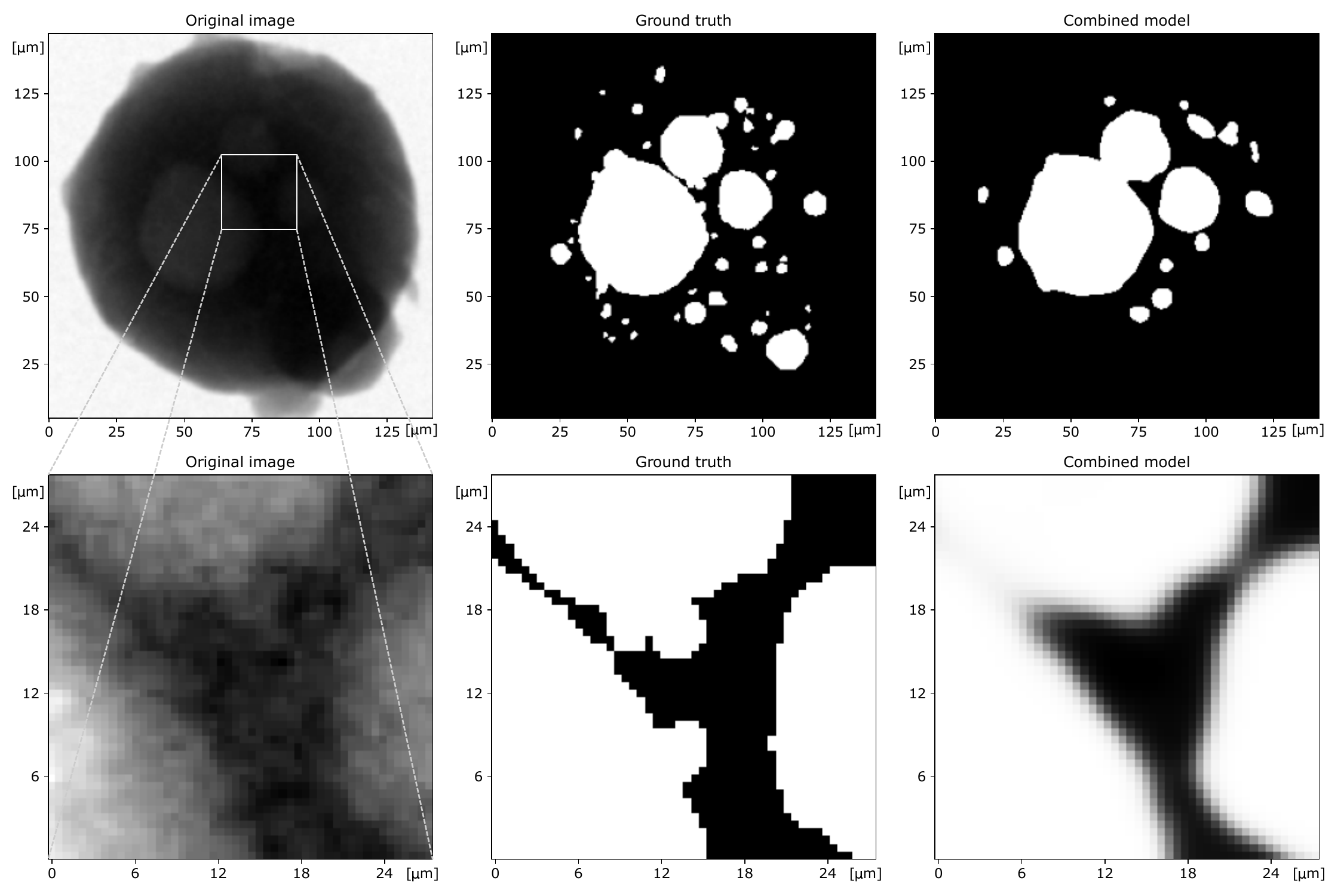}
    \caption{Segmentation output on test set.}
    \label{fig:set2results}
\end{figure*}

Next, the distributions of the test set pore predictions are investigated. Figure~\ref{fig:areadistreal} shows the distributions of the true and predicted pore radii in a violin plot, allowing for comparison between models. 
The figure clearly shows that the size of the smallest detectable pore is larger for the threshold-based models than for both, the ground truth and UNet-based models. We also see that the \textsc{Combined} model detects more small pores than the \textsc{UNet} model. In general, noisy images will make separation of nearby pores in the projections difficult and most methods will tend to overestimate the area of overlapping pores --- apparent in Figure~\ref{fig:set2results}.

For two nearly touching pores, the intensity in the separation can be difficult to distinguish from noisy variation inside or outside of a pore. This may cause the estimate of the number of pores or the projected area to be biased. For neural network models, loss functions may be designed to emphasize precision for some improvement, but best would likely be a specific pore-shape model, which is outside the scope of this article. Furthermore, UNet models particularly will tend to smooth the cusps at the boundary of intersecting pores (see again Figure~\ref{fig:set2results}). In this case, the total pore area will be overestimated.

Although this plot does not make a direct comparison between the pores detected and the ground truth, together with the observation from Figure~\ref{fig:error_real}, it supports the conclusion that while the \textsc{Combined} model is superior, it still underestimates the amount of small pores. 

A previous article \cite{Jaenisch2020} investigates the radiographic visibility limit of pores in metal powder for AM based on the data acquisition setup. Considering the similarity of the X-ray setup of \cite{Jaenisch2020} and this work, the smallest detectable pore size from radiographs was predicted to be at least 2~µm in radius in best case, e.g., by optimal experiment, optimal pore location, and optimal amount of pores. The smallest pore predictions are: \textsc{Local}: 5.8~µm, \textsc{Att.-Adj.}: 8.8~µm, \textsc{UNet}: 0.8~µm, \textsc{Combined}: 0.6~µm. Regardless of the smallest sizes, the models do not provide a guarantee on pore recognition. Instead, given the qualitative results, the probability for recognizing a pore increases rapidly with its size --- further investigation may quantify this relation.  Moreover, one has to take into account that a single radiographic view on a 3D object (particle) is not a completely sufficient description of the sizes and locations of pores (or any feature) found in a particle, e.g., pores may overlap, pores may be behind each other, pores may be not spherical but elongated and badly oriented or located, pores may interfere with surface roughness of the particle or satellites. Even if the smallest pores are not detected by the \textsc{Combined} model, the presented approach found most of the important pores with high accuracy, especially such pores which influence the AM process significantly, and hence the model is a good starting point for further optimizations and sensitivity tuning. 

\section{Conclusions}
To conclude, this work has looked at methods to detect and segment gas pores in feedstock powder for AM. The presence of these pores greatly affects the melting behaviour and porosity of the final AM product, and it is thus desirable to incorporate a porosity analysis during quality assurance. Classical image processing software, like Volume Graphics, Aviso or Dragonfly, is optimized for the analysis of 3D XCT data. But since complete XCT scans with a full rotation around the sample are time consuming, motivation exists to explore methods that allow for high throughput. By restricting our analysis to a single 2D radiographic projection image with non-overlapping powder particles, we can simulate possible future efficient setups.

The radiographic simulation tool \textit{aRTist} was successfully employed to generate an abundance of artificial radiographs in a controlled virtual environment for the pretraining of a UNet. However, the results show that real radiographs may contain complexities (e.g., surface roughness, form deviations, etc.) which are not captured by simple spherical synthetic particles and hence are tricky to simulate autonomously. When using artificially generated data from \textit{aRTist}, future virtual powder experiments need to focus on more realistic particle shapes, e.g., mimic shapes with surface-triangulated-latices (STL) to be imported into the virtual \textit{aRTist} environment. Real data affects the segmentation performance considerably which is apparent in Figure~\ref{fig:training_history} and further experiments from a preceding work \cite{andreas}. Modelling the data prior to a UNet forward pass shows considerably higher accuracy (increase of $11.4\%$ in F1-score). Here, a few different factors made UNet-based pore segmentation possible and generally successful: pretraining on synthetic data, making snugly fitting square image crops, and learning a signed distance map to synthesize artificial particles without pores ("ideal" for the given particle shape). To achieve even better segmentation results, other AI methods can be explored. In this space, a pixel to pixel generative adversarial model could capture other semantics than possible in the used UNet-like model --- and even learn the associated loss function (\cite{i2i}). Because of their scalable nature, and since this segmentation task was based on single 2D radiographs, these strategies appear promising in making high throughput porosity analysis of metal powder. 

The models derived in this study are material invariant. The investigation of other alloys requires the fine tuning of the radiography setup in order to achieve a sufficient contrast between bulk material and gas pore. If, for example, the bulk material is of lower density than the alloy used in this study ((Mn,Fe)$_2$(P,Si)), the contrast between bulk material and pore is reduced. This has to be compensated by changing the X-ray spectrum (here lowering the voltage) in order to increase the fraction of low-energetic photons being absorbed in the bulk material.
Hence, based on an optimal X-ray radiographic setup, the models presented here are not even limited to finding pores (low density) in metal particles (high density). Other material combinations and even a vice versa contrast situation (high-density inclusion in low-density particle) are feasible applications. 
As long as the radiographic setup allows to achieve a suitable contrast, the algorithms are capable to segment each type of defect with a given precision.

To approach more realistic in- or at-line quality control setups, a high throughput acquisition setup is necessary. Hence, a radiographic device for sizing and internal feature detection of scattered powder particles must be considered; and, as this work shows with an F1-score of $0.87$ during test segmentation, a single 2D radiograph could provide sufficient accuracy for many quality control and/or quality assurance use cases. By combining projections from several directions (e.g., separated by $90^o$ degrees), one could achieve an even more accurate measure of porosity and better detect pores which are small or close to the circular edge. High throughput requires short exposure times, which subsequently results in more image noise. To compensate for short exposure times, a higher photon flux (i.e., larger X-ray source power) is necessary, to the detriment of image resolution. Possible workarounds here can be the use of \textit{time-delay integration} (TDI) scans, similar to imaging sensor readout architectures~\cite{Holdsworth1990, Lepage2009}. This will allow for short exposure times in each frame but yield reasonable CNR after spatially corrected pixel integration.
On the other hand, this scanning technique could also be used to synthetically increase the image resolution like shown in \cite{Farsiu2004,  Fruchter2002}.

A related topic would be to annotate powder radiographs with measures of feedstock flowability and subsequently train suitable ML models for inferring flowability.
This would enable determination of both porosity and flowability of feedstock powder from, e.g., a simple 2D radiographic inspection device in a conveyor belt-style arrangement.

\section{Data availability}
The raw data required to reproduce these findings are available to download from \textit{Zenodo}~\cite{Schumacher2022}. The processed data required to reproduce these findings are available to download from \textit{Zenodo}~\cite{Schumacher2022}.

\printcredits

\bibliographystyle{unsrtnat}

\bibliography{Bibliography_AMPoreDetection} 

\newpage\onecolumn
\appendix
\section{Large render of the 3D powder particle volumes; hue encodes equivalent pore diameter}\label{app:render}
{
  \centering
   \includegraphics[angle=0, width=\linewidth]{stitched_3D.png}
}

\section{Flowchart model overview }\label{fig:overview}
\begin{figure*}[pos=h!]
    \centering
    \includegraphics[width=1\linewidth]{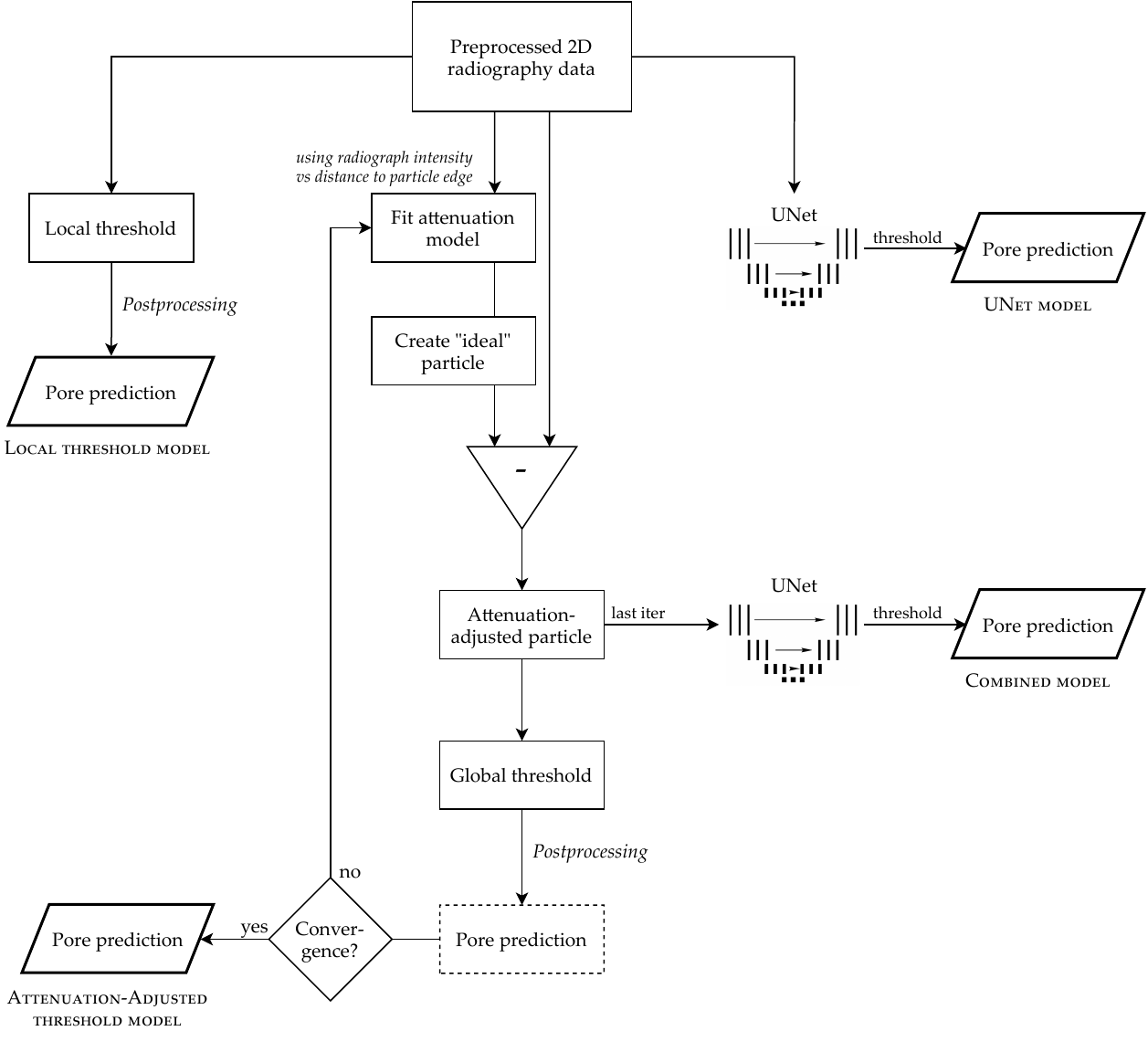}
\end{figure*}\vspace{-0.5cm}
Overview and process pipeline of the four models. Iteration count for convergence of the attenuation-adjusted model is estimated empirically. The two UNet models are pretrained on raw \textit{aRTist} artificial data and attenuation-adjusted artificial data, respectively.

\section{Graphical abstract}
{\centering
\includegraphics[width=\linewidth]{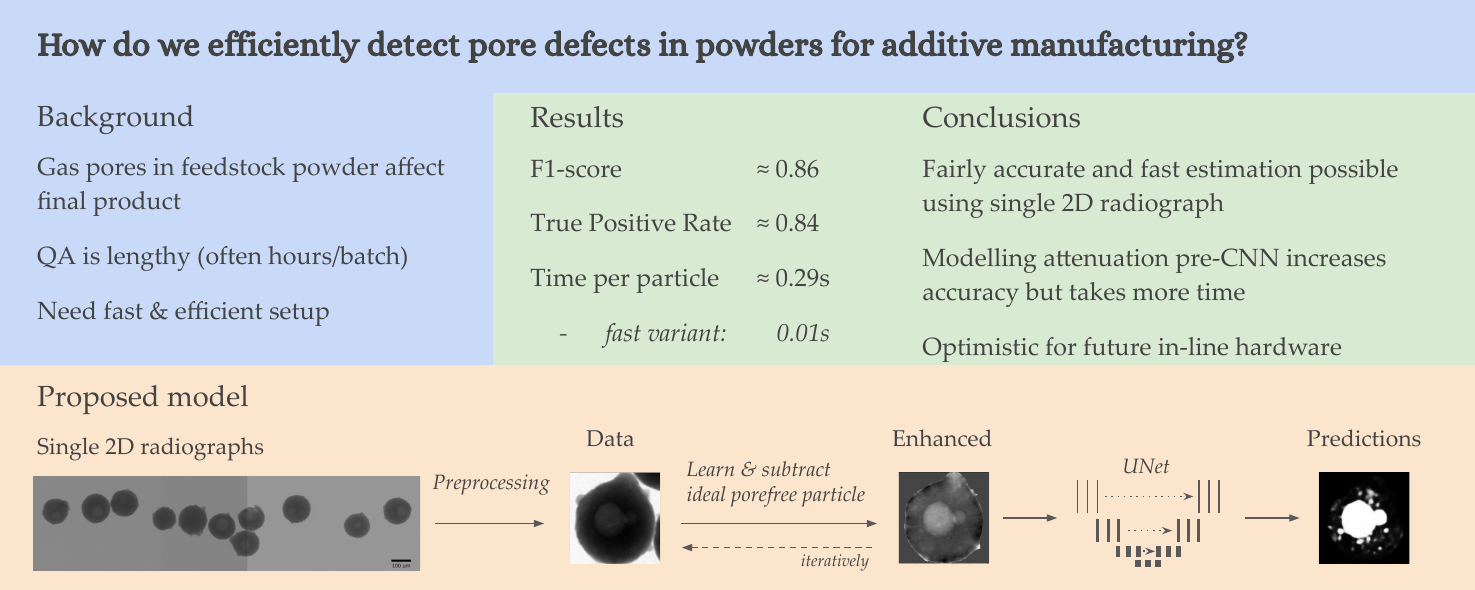} 
}

\bio{}
\endbio

\end{document}